\documentclass{article}

\usepackage[preprint,nonatbib]{neurips_2024}

\usepackage[utf8]{inputenc} % allow utf-8 input
\usepackage[T1]{fontenc}    % use 8-bit T1 fonts
\usepackage{hyperref}       % hyperlinks
\usepackage{url}            % simple URL typesetting
\usepackage{booktabs}       % professional-quality tables
\usepackage{amsfonts}       % blackboard math symbols
\usepackage{nicefrac}       % compact symbols for 1/2, etc.
\usepackage{microtype}      % microtypography
\usepackage{xcolor}         % colors
\usepackage{graphicx}  % Required for including images
\usepackage{caption}   % Allows for customized captions
\usepackage{multirow}
\usepackage{colortbl}
\usepackage{tabularx}
\usepackage{array}
\usepackage{subcaption}
\usepackage{listings}
\usepackage{xcolor}
\usepackage{float}
\usepackage{amsmath}
\usepackage[numbers]{natbib}

\DeclareFontShape{T1}{ptm}{m}{scit}{<->ssub*ptm/m/it}{}

\title{Self-Supervised Anomaly Detection in the Wild: Favor Joint Embeddings Methods}

\author{
  Daniel Otero\thanks{Equal contribution} \\
  EXIT83\thanks{EXIT83 LLC Consulting Services, Seattle, United States} \\
  Medellin, Colombia \\
  \texttt{daniel@exit83.com} \\
  \And
  Rafael Mateus\footnotemark[1] \\
  EXIT83\footnotemark[2] \\
  Medellin, Colombia \\
  \texttt{rafael@exit83.com} \\
  \And
  Randall Balestriero \\
  Brown University  \\
  Providence, RI, USA \\
  \texttt{rbalestr@brown.edu} \\
}

\begin{document}

\maketitle

\begin{abstract}

Accurate anomaly detection is critical in vision-based infrastructure inspection, where it helps prevent costly failures and enhances safety. Self-Supervised Learning (SSL) offers a promising approach by learning robust representations from unlabeled data. However, its application in anomaly detection remains underexplored. This paper addresses this gap by providing a comprehensive evaluation of SSL methods for real-world anomaly detection, focusing on sewer infrastructure. Using the Sewer-ML dataset, we evaluate lightweight models such as ViT-Tiny and ResNet-18 across SSL frameworks, including BYOL, Barlow Twins, SimCLR, DINO, and MAE, under varying class imbalance levels. Through 250 experiments, we rigorously assess the performance of these SSL methods to ensure a robust and comprehensive evaluation. Our findings highlight the superiority of joint-embedding methods like SimCLR and Barlow Twins over reconstruction-based approaches such as MAE, which struggle to maintain performance under class imbalance. Furthermore, we find that the SSL model choice is more critical than the backbone architecture. Additionally, we emphasize the need for better label-free assessments of SSL representations, as current methods like RankMe fail to adequately evaluate representation quality, making cross-validation without labels infeasible. Despite the remaining performance gap between SSL and supervised models, these findings highlight the potential of SSL to enhance anomaly detection, paving the way for further research in this underexplored area of SSL applications.

\end{abstract}

%%%%%%%%%%%%%%%%%%%%%%%%%%%%%%%%%%%%%%%%%%%%%%%%%%%%%%%%%%%%%%
\section{Introduction}

Self-Supervised Learning (SSL) is a machine learning paradigm where models are trained on unlabeled data by creating surrogate labels through pretext tasks that exploit inherent structures or patterns within the data. As a result, this approach enables learning meaningful representations that can be applied to various downstream tasks without the need for explicit manual labeling \citep{balestriero2023acookbook}. Because of this, SSL is particularly advantageous for semi-supervised anomaly detection problems where obtaining labeled data is costly, labor-intensive, impossible, or undesirable \citep{akoglu2023ssltackling}. Despite these benefits, anomaly detection is frequently underrepresented in SSL research, with recent large-scale ablation studies often neglecting its inclusion in benchmarking \citep{goldblum2023battle, al-tahan2024unibench}. In fact, common benchmarks  such as ImageNet \citep{deng2009imagenet} and CIFAR \citep{krizhevsky2009learning}, are object-centric and do not accurately reflect the complexity of real-world environments, where images are more diverse and less structured \citep{goyal2021selfsupervised}.

Closer to the anomaly detection scenario, recent works have started to stress-test SSL on more realistic scenarios with uncurated data. Albeit still employing a classification task evaluation, it has been shown that SSL may be sensitive to the quality of the data and in particular to balance representations of the features to be learned \citep{vo2024automatic, assran2023thehiddenuniform}. Clearly, such warning seems to go against the ability of SSL to solve anomaly detection as such scenario assumes by nature that the model can learn the anomaly features from very few samples. 

This gap in evaluation coupled with those recent sensitivity studies of SSL beg the following question:

{\em Can SSL successfully learn representations on highly imbalance datasets and capture salient features to solve anomaly detection tasks?}

\begin{figure}[t!]
    \centering
    \begin{minipage}{0.29\linewidth}
    \includegraphics[width=\linewidth]{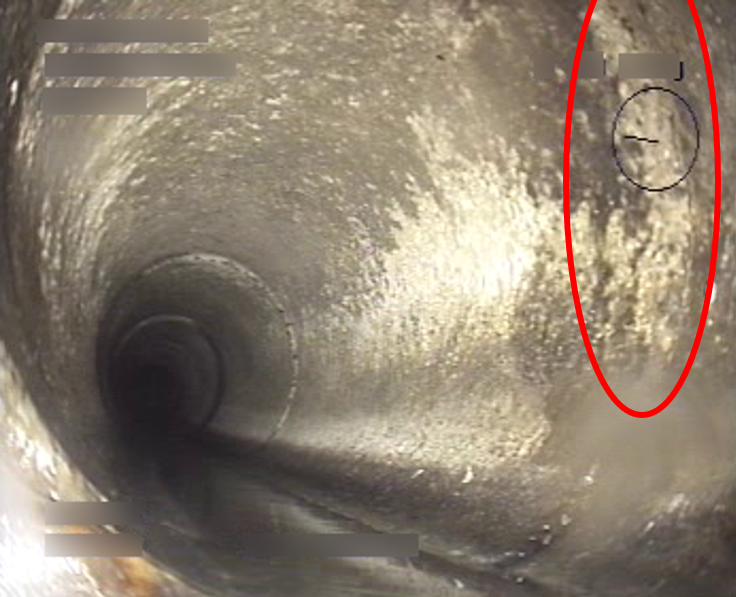}
    \end{minipage}
    \begin{minipage}{0.29\linewidth}
    \includegraphics[width=\linewidth]{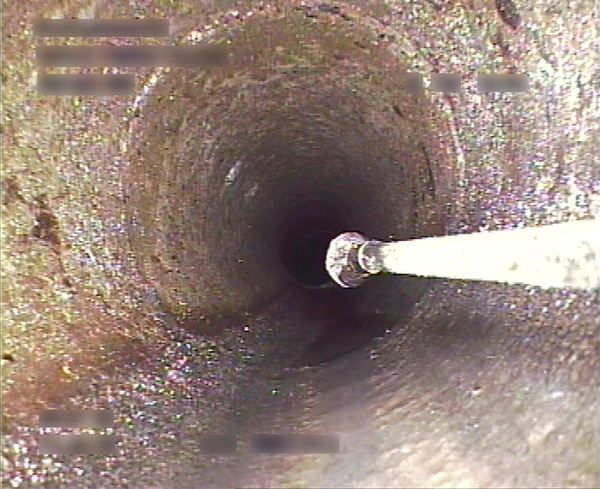}
    \end{minipage}
    \begin{minipage}{0.38\linewidth}
    \caption{\small Samples from the Sewer-ML dataset \citep{harum2021sewer}. The red circle highlights a defect associated with lateral reinstatement cuts, where improper cutting or misalignment can cause issues such as blockages, leaks, or structural weakness in the sewer system (for further details on defect types and image samples used for training and validation, refer to Appendix~\ref{app:train-val-image-samples}).}
    \label{fig:sewerml-images}
    \end{minipage}
\end{figure}

To scientifically approach that question, we propose the use of the Sewer-ML dataset \citep{harum2021sewer}, a sewerage infrastructure dataset that contains 1.3 million images captured from video inspections, featuring 17 different defect classes. Figure~\ref{fig:sewerml-images} presents examples of defect and non-defect images taken from Sewer-ML. 
To determine the robustness of self-supervised learning in handling class imbalances, we conducted 250 binary classification experiments where we systematically vary the proportion of defect samples in the train and validation datasets as 1\%, 2\%, 5\%, and 15\%.

We summarize our key findings below:
\begin{itemize}
    \item {\bf Superiority of joint-embedding methods over reconstruction methods:} we observe that MAE has difficulties to maintain performance with varying class imbalance as opposed to methods like SimCLR or BarlowTwins that compete with supervised baselines.
    \item {\bf Impact of SSL methodology over backbone architecture:} the choice of backbone architecture (Resnet or Transformer) is not critical compared to the choice of SSL method.
    \item {\bf Need for better label-free assessment of SSL representations:} our findings indicate that methods such as RankMe fail to assess the richness of SSL representations making cross-validation without labels currently impossible on such task and dataset.
\end{itemize}
The code for this research will be open-sourced and is available at \href{https://drive.google.com/file/d/195NxZN3-4tT9sOJCiUD3c6OkauZsQ5UA/view?usp=sharing}{Anomaly-Detection-In-The Wild\_code.zip}. 

\section{Controlled evaluation of self-supervised anomaly detection}

We will first describe our methodology in detail and then provide results and discussions at the end of this section.

\subsection{Methodology}

Prior to conducting the SSL ablation study, an initial hyperparameter search was performed to optimize the data augmentation pipeline for the dataset. The focus was placed on tuning image resolution and augmentation settings to achieve a balance between model performance and computational efficiency. Detailed results and configurations are provided in Appendix~\ref{sec:res_and_da_ablation}.

{\bf Methods and models.}~Our study conducts an ablation analysis on anomaly detection using self-supervised learning methods, with a particular emphasis on their robustness to distribution imbalances. We primarily focus on joint-embedding architectures—specifically Barlow Twins, SimCLR, BYOL, and DINO \citep{zbontar2021barlow, chen2020asimple, grill2020bootstrap, caron2021emerging}—which aim to learn an embedding space by aligning representations of different augmented views of the same input while avoiding collapse. To provide a comparative perspective, we also evaluate Masked Autoencoders (MAE) \citep{he2022masked}, a self-supervised approach that reconstructs missing parts of the input data. For our backbone architectures, we use lightweight models such as ViT-Tiny \citep{dosovitskiy2020image} and ResNet-18 \citep{he2016deep}. Please refer to Appendix~\ref{sec:ssl_config} to inquire about specific training hyperparameters used for each method. 

{\bf Controlled dataset imbalance.}~To determine the robustness of self-supervised learning in handling class imbalances, we conduct binary classification experiments with the Sewer-ML dataset \citep{harum2021sewer}, where we systematically vary the proportion of defect samples in the dataset. Specifically, we assess the performance of these methods under different levels of class imbalance—defect sample proportions of 1\%, 2\%, 5\%, and 15\%. As a binary setting, Sewer-ML has an approximately balanced distribution by design-45\% and 47\% defect proportion on the train and validation sets, respectively. We seek to assess how well self-supervised learning approaches can learn from imbalanced distributions and to evaluate their effectiveness on datasets with varying levels of class imbalance. We apply each of the specified defect proportions to both the training and testing datasets, allowing us to evaluate model performance across all combinations of class imbalance scenarios.

{\bf Evaluation metrics.}~In addition to reporting performance based solely on the F1 score, we also evaluate our models using the metrics proposed by the benchmark. This includes a weighted F2 metric (F2$_{CIW}$) for identifying defects and a conventional F1 score (F1$_{Normal}$) for non-defect classifications \citep{harum2021sewer}. The F2 metric's weights are assigned to each defect category according to their economic significance. Additionally, they encourage the use of an F2 score to give greater emphasis to recall over precision, recognizing that overlooking a defect incurs a higher economic cost than the occurrence of a false positive.

%%%%%%%%%%%%%%%%%%%%%%%%%%%%%%%%%%%%%%%%%%%%%%%%%%%%%%%%%%%%%%
\subsection{Results}

We present in Table~\ref{tab:val_trend_metrics} the performance of various SSL models compared to their supervised counterparts. Training was conducted on imbalanced data, while validation adhered to a balanced set, aligning with benchmark standards. Although previous studies have noted that CNN-based architectures like ResNet-18 often outperform vision transformers (ViTs) \citep{goldblum2023battle}, our results do not indicate a significant difference in performance between these backbones architectures.

{\bf Robustness of joint-embedding methods.}~When examining individual SSL methods, SimCLR achieves the best overall results especially when paired with ResNet-18 and defect proportion is higher than 5\%. When working with stronger distribution imbalances, particularly with 1\%, BYOL and Resnet-18 have the best results. This might be due to the training dynamics behind these methods, in situations with moderate imbalance (like 5\%), the contrastive approach of SimCLR can still adequately separate the minority class from the majority, leveraging the discriminative power of the contrastive loss. This can be counter-productive when working with extremely imbalance levels. BYOL avoids this issue by not requiring explicit negative sampling, allowing it to maintain more consistent performance even when the class imbalance becomes more extreme. Thus, while SimCLR thrives with moderate imbalance, BYOL proves more resilient in handling extreme class disparities.
Another interesting insight is Barlow Twins’ competitive performance under extreme imbalance. Its use of redundancy reduction loss, which both maximizes similarity between augmented views and decorrelates the learned representations, may help avoid the pitfalls of overfitting to the dominant class. DINO, however, shows a unique trend, performing poorly in 1\% and 2\% settings with ResNet-18, but not with ViT-Tiny. This could be attributed to DINO’s focus on global feature learning through knowledge distillation, which aligns better with ViT’s global attention approach.

{\bf Failure of reconstruction-based methods.}~Finally, the Masked Autoencoder (MAE) delivers the weakest performance across all scenarios, especially at higher imbalance levels (e.g., 45\%). MAE’s reliance on reconstructing multiple classes might introduce noise that hampers its ability to generalize under severe imbalance. In highly imbalanced data, MAE could face challenges in differentiating between common and rare classes, possibly due to the reconstruction bias that favors the majority class. This might also be tied to the model’s neural capacity, which can struggle to produce robust representations when not sufficiently overparameterized.

Full tables with training and validation results are in Appendix~\ref{sec:full_balanced_tables}. Detailed analysis of each model's performance on imbalanced test distributions is provided in Appendix~\ref{sec:imbalanced_validation_set}. SSL monitoring metrics including RankMe \citep{garrido2023rankme} and the mean and standard deviation of features and embeddings for DINO, BYOL, and SimCLR—are discussed in Appendix~\ref{app:ssl_monitor_metrics}.

\begin{table}[t]
\caption{Validation metrics for each method and architecture across different imbalance levels. The percentages represent the imbalance levels applied during training, while the validation data proportion remained unchanged.}
\label{tab:val_trend_metrics}
\vskip 0.15in
\begin{center}
\begin{small}
\begin{sc}
\renewcommand{\arraystretch}{0.8}
\resizebox{\columnwidth}{!}{%
\begin{tabular}{llrrrrrrrrrr}
\toprule
     & & \multicolumn{5}{c}{\textbf{ResNet-18}} & \multicolumn{5}{c}{\textbf{ViT-Tiny}} \\
     & & \textbf{1\%} & \textbf{2\%} & \textbf{5\%} & \textbf{15\%} & \textbf{45\%} & \textbf{1\%} & \textbf{2\%} & \textbf{5\%} & \textbf{15\%} & \textbf{45\%}\\
\midrule
\multirow{6}{*}{F1$\uparrow$} & \textit{Supervised}  & \cellcolor{gray!20}0.831 & \cellcolor{gray!20}0.847 & \cellcolor{gray!20}0.865 & \cellcolor{gray!20}0.882 & \cellcolor{gray!20}0.898 & \cellcolor{gray!20}0.768 & \cellcolor{gray!20}0.785 & \cellcolor{gray!20}0.831 & \cellcolor{gray!20}0.858 & \cellcolor{gray!20}0.878 \\
\cmidrule(lr){2-12}
     & \textit{BYOL}  & 0.764 & 0.768 & 0.774 & 0.783 & 0.792 & 0.676 & 0.738 & 0.752 & 0.757 & 0.730 \\
     & \textit{Barlow Twins}  & 0.768 & 0.775 & 0.789 & 0.796 & 0.803 & 0.682 & 0.738 & 0.747 & 0.769 & 0.772 \\
     & \textit{DINO}  & 0.049 & 0.032 & 0.673 & 0.719 & 0.727 & 0.691 & 0.683 & 0.553 & 0.704 & 0.719 \\
     & \textit{MAE}  & -- & -- & -- & -- & -- & 0.513 & 0.636 & 0.570 & 0.575 & 0.024 \\
     & \textit{SimCLR}  & 0.740 & 0.778 & 0.796 & 0.807 & 0.814 & 0.762 & 0.768 & 0.771 & 0.778 & 0.786 \\
\midrule
\multirow{6}{*}{F2$_{CIW}\uparrow$} & \textit{Supervised}  & \cellcolor{gray!20}0.838 & \cellcolor{gray!20}0.849 & \cellcolor{gray!20}0.880 & \cellcolor{gray!20}0.904 & \cellcolor{gray!20}0.879 & \cellcolor{gray!20}0.837 & \cellcolor{gray!20}0.812 & \cellcolor{gray!20}0.831 & \cellcolor{gray!20}0.870 & \cellcolor{gray!20}0.881 \\
\cmidrule(lr){2-12}
     & \textit{BYOL}  & 0.837 & 0.844 & 0.819 & 0.825 & 0.785 & 0.680 & 0.768 & 0.760 & 0.781 & 0.725 \\
     & \textit{Barlow Twins}  & 0.817 & 0.805 & 0.817 & 0.822 & 0.780 & 0.704 & 0.749 & 0.786 & 0.784 & 0.758 \\
     & \textit{DINO}  & 0.566 & 0.402 & 0.712 & 0.777 & 0.752 & 0.733 & 0.736 & 0.690 & 0.724 & 0.695 \\
     & \textit{MAE}  & -- & -- & -- & -- & -- & 0.629 & 0.640 & 0.704 & 0.702 & 0.009 \\
     & \textit{SimCLR}  & 0.775 & 0.813 & 0.826 & 0.822 & 0.813 & 0.766 & 0.795 & 0.798 & 0.801 & 0.788 \\
\midrule
\multirow{6}{*}{F1$_{Normal}\uparrow$} & \textit{Supervised}  & \cellcolor{gray!20}0.850 & \cellcolor{gray!20}0.861 & \cellcolor{gray!20}0.879 & \cellcolor{gray!20}0.894 & \cellcolor{gray!20}0.908 & \cellcolor{gray!20}0.774 & \cellcolor{gray!20}0.794 & \cellcolor{gray!20}0.844 & \cellcolor{gray!20}0.869 & \cellcolor{gray!20}0.890 \\
\cmidrule(lr){2-12}
     & \textit{BYOL}  & 0.758 & 0.759 & 0.783 & 0.789 & 0.818 & 0.719 & 0.757 & 0.778 & 0.771 & 0.781 \\
     & \textit{Barlow Twins}  & 0.782 & 0.796 & 0.799 & 0.807 & 0.828 & 0.716 & 0.760 & 0.766 & 0.785 & 0.806 \\
     & \textit{DINO}  & 0.689 & 0.686 & 0.700 & 0.730 & 0.763 & 0.655 & 0.688 & 0.419 & 0.727 & 0.759 \\
     & \textit{MAE}  & -- & -- & -- & -- & -- & 0.471 & 0.350 & 0.553 & 0.389 & 0.677 \\
     & \textit{SimCLR}  & 0.752 & 0.793 & 0.806 & 0.818 & 0.836 & 0.782 & 0.781 & 0.786 & 0.785 & 0.810 \\
\bottomrule
\end{tabular}%
}
\end{sc}
\end{small}
\end{center}
\vskip -0.1in
\end{table}

\begin{figure}[ht]
    \centering
    \includegraphics[width=\linewidth]{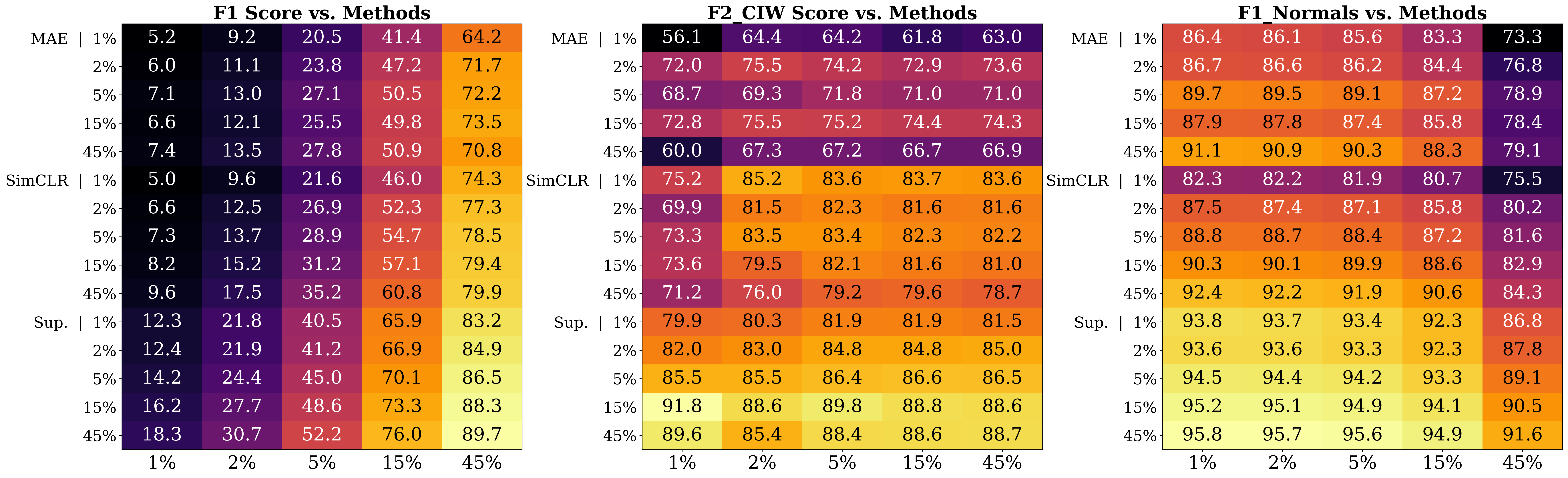}
    \caption{ResNet-18 validation performance heatmaps across imbalance levels. The x-axis represents the imbalance levels in the validation set, while the y-axis indicates the method and the imbalance level used during training.}
    \label{fig:heatmap-resnet}
\end{figure}

\textbf{Performance Trends on Imbalanced Validation Sets.}~When evaluating the models on increasingly imbalanced validation sets (Figure~\ref{fig:heatmap-resnet}), there is a clear and consistent decline in performance across most methods, as indicated by the F1 scores. As the validation set imbalance becomes more severe (e.g., at 1\% and 2\% imbalance), no method is able to sustain strong performance, particularly in terms of accurately identifying the minority class. On the other hand, the F2 score remains relatively stable, likely due to the use of a weighted loss function based on defect proportions.

%%%%%%%%%%%%%%%%%%%%%%%%%%%%%%%%%%%%%%%%%%%%%%%%%%%%%%%%%%%%%
\section{Conclusions and discussion}

Our study indicates that self-supervised learning (SSL) is effective for anomaly detection and remains robust even when facing significant distribution imbalances. We also find that the choice of backbone architecture is not the most critical factor in model performance, as neither ViT-Tiny nor ResNet-18 consistently outperforms the other across all cases. In contrast, the selection of the SSL methodology significantly impacts performance, with substantial variations observed among different SSL model families. Therefore, for practitioners, choosing the appropriate SSL family is more crucial than selecting a specific backbone architecture.
Furthermore, there is a pressing need to accurately measure the quality of representations produced by SSL models. Our findings indicate that the RankMe metric is ineffective for this purpose; it aims to assess the richness of representations but fails to correlate with actual performance. As shown in Appendix~\ref{app:ssl_monitor_metrics}, there is no correlation between performance and RankMe metrics, underscoring the necessity for better methods of evaluating representations quality.

%%%%%%%%%%%% REFERENCES %%%%%%%%%%%%%%%%%%%%%%%%%%%%%%%%%

\bibliographystyle{plain}
\bibliography{references}

%%%%%%%%%%%%%% APPENDIX %%%%%%%%%%%%%%%%%%%%%%%%%%%%%%%%%%%

\appendix
\newpage

\section{Images resolution and data augmentations ablation}
\label{sec:res_and_da_ablation}

For the resolution, we trained both architectures in a supervised fashion, using no data augmentations other than resizing. As seen in Table~\ref{tab:resolution_ablation}, the resolution ablation study reveals a clear trend where increasing the image size leads to better model performance, as measured by validation F1 scores. However, this improvement comes at the cost of increased training time. Notably, the best trade-off between performance and runtime was achieved at a resolution of 224x224.

\begin{table}[H]
\centering
\caption{Best validation F1 scores for different image sizes and architectures.}
\begin{tabular}{lcccccc}
\toprule
 & \textbf{64} & \textbf{128} & \textbf{224} & \textbf{384} & \textbf{512} \\
\midrule
ViT-Tiny   & 0.866 & 0.885 & 0.894 & 0.897 & 0.90 \\
\midrule
ResNet-18  & 0.877 & 0.892 & 0.901 & 0.907 & 0.909 \\
\midrule
Time/Epoch & 28m & 31m & 37m & 52m & 69m \\
\bottomrule
\end{tabular}
\label{tab:resolution_ablation}
\end{table}

\begin{figure}[htbp]
    \centering
    \includegraphics[width=0.8\linewidth]{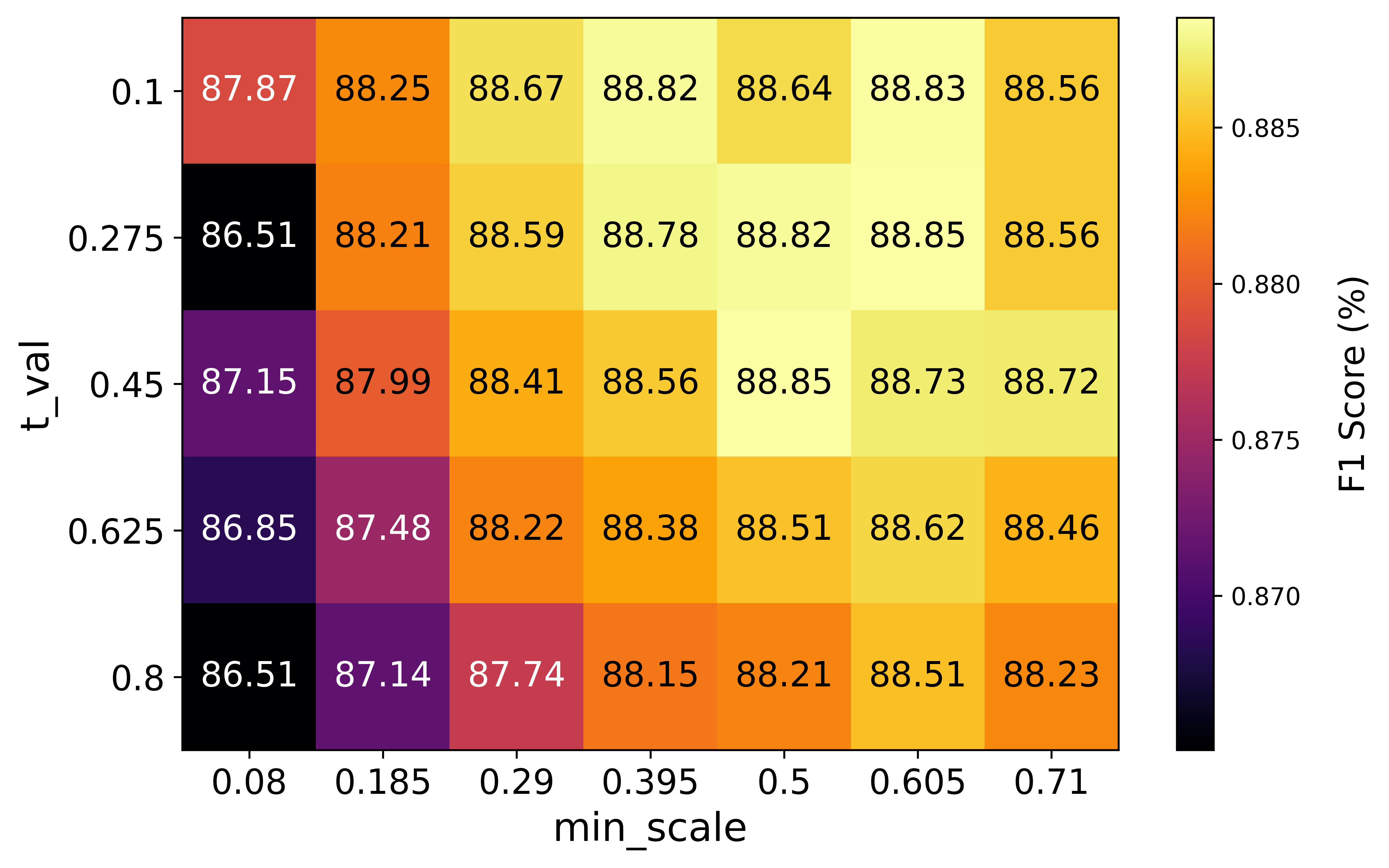}
    \caption{Val F1 Score by \texttt{min\_scale} and \texttt{t\_val}. This heatmap shows the performance variation in terms of F1 score, demonstrating the interaction between these two hyperparameters.}
    \label{fig:heatmap}
\end{figure}

For the data augmentations, we explored how changes in color jitter and random cropping impacted model performance. To maintain computational efficiency, we used 25\% of the dataset, running experiments with the ResNet-18 architecture. We adjusted the \texttt{ColorJitter} parameters using a variable \texttt{t\_val} that ranged from 0.1 to 0.8. Specifically, brightness, contrast, and saturation were each set to \texttt{t\_val}, while hue was set to half of that value (\text{hue} = \texttt{t\_val} / 2). Furthermore, the minimum scale for \texttt{RandomResizedCrop} was varied from 0.08 to 0.71. As illustrated in Figure~\ref{fig:heatmap}, the best performance, measured by F1 Score, was obtained when \texttt{t\_val} was around 0.1 and 0.45 and when \texttt{min\_scale} was around 0.395 and 0.605. Due to the small performance difference, and looking to introduce stronger augmentations when using SSL methodologies, we decided to use \texttt{min\_scale} = 0.395 and \texttt{t\_val} = 0.275 for further ablations.  

\section{SSL configurations}
\label{sec:ssl_config}

Training was conducted using PyTorch 2.0.2 \citep{paszke2019pytorch} on a SLURM cluster equipped with NVIDIA L4 GPUs. In total, we performed 250 experiments, comprising 74 training runs and 184 validation tests. The experiments were conducted over 45 epochs varying the optimizer and its corresponding parameters depending on the method at hand. A linear warmup was applied for the first 10 epochs, followed by a cosine scheduler with no restarts. The base and final decay rates (\(\tau\)) were 0.996 and 0.999, respectively, with a minimum learning rate of \(1 \times 10^{-6}\).

During pre-training, a linear probe was trained, and the reported results reflect its final performance. For all experiments, the linear probe was trained using a learning rate of 0.001, with a momentum of 0.9 and no weight decay applied. However, for MAE, a reduced learning rate of 0.0001 was necessary to prevent numerical issues caused by 16-bit mixed precision. Similarly, DINO was trained using 32-bit precision due to instability on imbalanced cases. 

Moreover, due to the imbalanced nature of the problem, we use a binary cross entropy loss with positive weights. The positive weights are assigned using the inverse of the defect proportion, a method commonly referred to as inverse class frequency weighting. Finally, every SSL method was trained on two augmented views of the same image and no multicrop was used. 

To ensure reproducibility, the remainder of the section contains the specific configurations adopted for each method. 

\subsection{BYOL}

\begin{lstlisting}
optimization:
  optimizer: lars
  batch_size: 256
  lr: 0.2
  momentum: 0.9
  weight_decay: 1.5e-06
  exclude_bias_and_norm: true
  warmup_start_lr: 0.1
ssl_settings:
  proj_hidden_dim: 2048
  pred_hidden_dim: 2048
  proj_output_dim: 256
  normalize_projector: false
\end{lstlisting}

\subsection{DINO}

\begin{lstlisting}
optimization:
  optimizer: adamw
  batch_size: 128
  lr: 0.0005
  momentum: 0.9
  weight_decay: 0.0001
  warmup_start_lr: 3.0e-05
ssl_settings:
  proj_hidden_dim: 2048
  proj_output_dim: 256
  num_prototypes: 8192
  clip_grad: true
  freeze_prototyper: 1
  use_bn_in_head: false
  norm_prototyper: false
  student_temperature: 0.1
  warmup_teacher_temperature: 0.04
  teacher_temperature: 0.07
  warmup_temperature_epochs: 15
\end{lstlisting}

\subsection{MAE}

\begin{lstlisting}
optimization:
  optimizer: adamw
  batch_size: 256
  lr: 0.001
  momentum: 0.9
  weight_decay: 0.05
  warmup_start_lr: 0.0005
ssl_settings:
  mask_ratio: 0.75
  norm_pix_loss: true
  decoder_embed_dim: 192
  decoder_depth: 12
  decoder_num_heads: 12
\end{lstlisting}

\subsection{SimCLR}

\begin{lstlisting}
optimization:
  optimizer: lars
  batch_size: 256
  lr: 0.3
  momentum: 0.9
  weight_decay: 1.0e-06
  warmup_start_lr: 0.15
ssl_settings:
  proj_hidden_dim: 2048
  proj_output_dim: 256
  temperature: 0.2
\end{lstlisting}

\subsection{Barlow Twins}

\begin{lstlisting}
optimization:
  optimizer: lars
  batch_size: 256
  lr: 0.2
  momentum: 0.9
  weight_decay: 1.5e-06
  warmup_start_lr: 0.1
ssl_settings:
  proj_hidden_dim: 2048
  proj_output_dim: 256
  lambda: 0.0051
  scale_loss: 0.024
\end{lstlisting}

\subsection{Supervised}

\begin{lstlisting}
optimization:
  optimizer: adamw
  batch_size: 256
  lr: 0.0005
  momentum: 0.9
  weight_decay: 0.0001
  warmup_start_lr: 3.0e-05
\end{lstlisting}

\section{Complementary tables}

\begin{figure}[H]
    \centering
    \includegraphics[width=\linewidth]{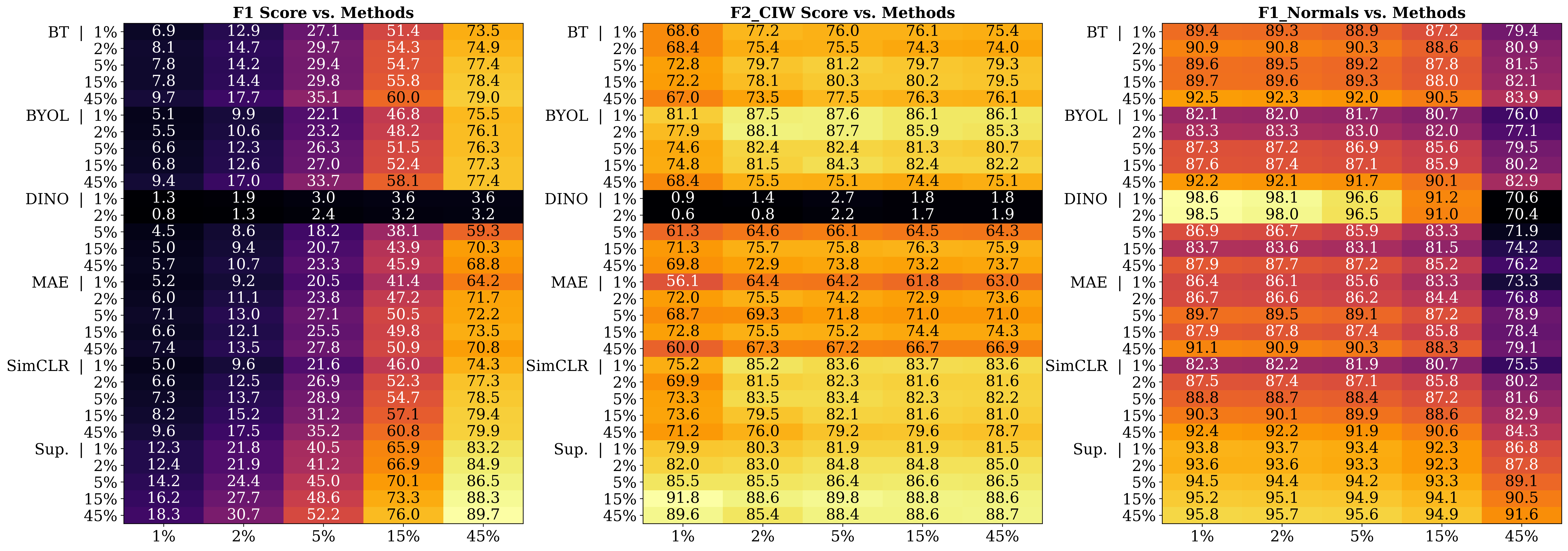}
    \caption{ResNet-18's F1, F2$_{CIW}$, and F1$_{Normal}$ validation score heatmaps across imbalance levels. The x-axis represents the imbalance levels in the validation set, while the y-axis indicates the method and the imbalance level used during training.}
    \label{fig:complete_main_heatmap}
\end{figure}

When evaluating SSL methods trained and tested on various class imbalances, several important observations emerged (refer to Figure~\ref{fig:complete_main_heatmap}). DINO, when trained on low defect proportions (1\% and 2\%), struggled significantly in identifying defects, despite performing well at detecting non-defects, suggesting a limitation in handling extreme class imbalances. Additionally, the supervised baseline consistently outperformed all SSL methods across every class imbalance, with the performance gap widening as the class imbalance in the validation set increased. This highlights the need for further exploration of SSL methods in real-world anomaly detection scenarios. Interestingly, Barlow Twins demonstrated the strongest resilience in highly imbalanced settings, particularly when both training and evaluation involved significant class imbalances. Specifically, Barlow Twins variants trained on 1\% and 2\% defect proportions achieved the best performance among SSL methods under these conditions, underscoring its robustness in scenarios with severe class imbalances.

\begin{figure}[H]
    \centering
    \includegraphics[width=\linewidth]{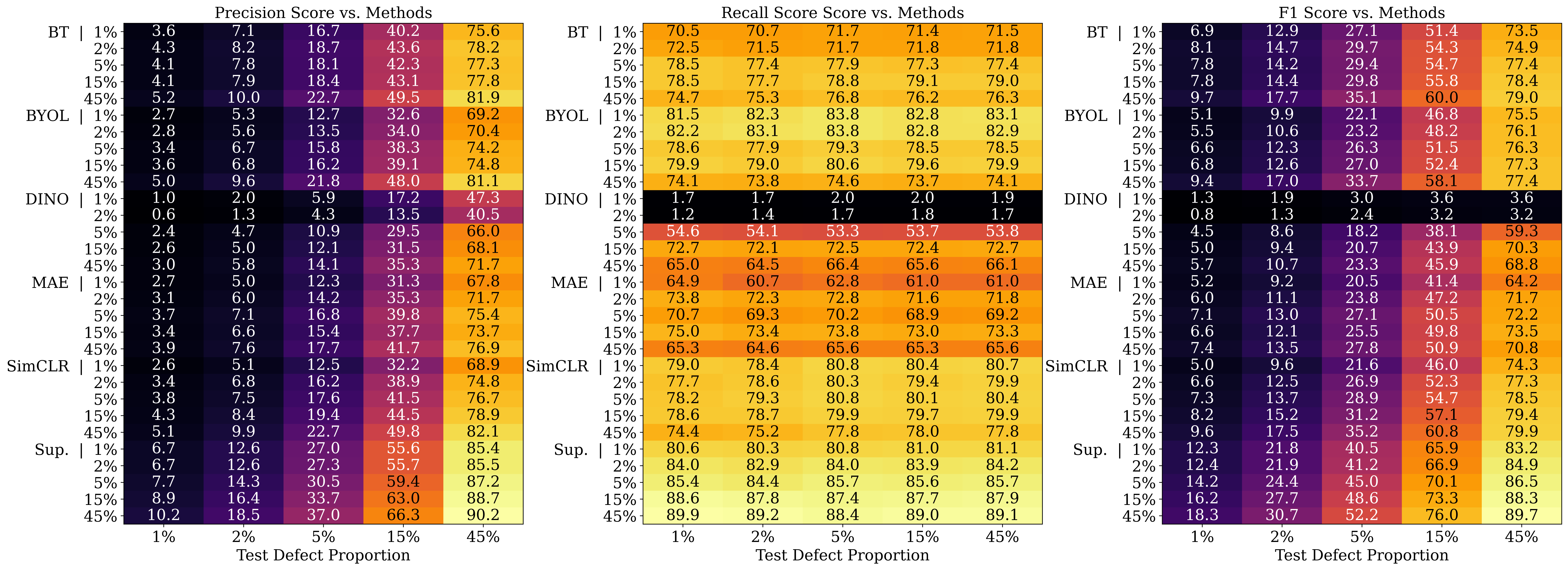}
    \caption{ResNet-18's precision, recall, and f1 validation score heatmaps across imbalance levels. The x-axis represents the imbalance levels in the validation set, while the y-axis indicates the method and the imbalance level used during training.}
    \label{fig:complete_sec_heatmap}
\end{figure}

When evaluating precision across varying validation imbalances, a degradation trend is evident as class imbalance increases (refer to Figure~\ref{fig:complete_sec_heatmap}). Notably, BYOL, when trained on a nearly balanced setting (45\% defect proportion), shows a larger performance gap compared to its 15\% counterpart, indicating the potential influence of training balance. Meanwhile, the recall score remains consistently strong across all validation imbalance proportions, likely due to the weighted loss choice, demonstrating solid recall performance even under significant class imbalances.

\subsection{Imbalanced validation setting}
\label{sec:imbalanced_validation_set}

\subsection{Balanced validation setting}
\label{sec:full_balanced_tables}

Overall, the training performance across the models show a clear linear degradation as the proportion of defects in the dataset decreases. This decline is primarily reflected in the precision scores, which drop significantly at lower defect proportions, such as 1\% and 2\%. The decreasing precision indicates that as fewer defects are present in the training set, the models fail to generalize properly, as the limited number of defect examples hinders their ability to accurately distinguish between defective and non-defective instances, leading to an increase in false positives. However, this trend is less evident in the validation set, where the precision remains relatively stable. This is due to the balanced nature of the validation set. On the other hand, recall remains fairly consistent across both the training and validation sets, even as the defect proportion decreases. This behavior can be explained by the use of a weighted loss function, which adjusts weights based on defect proportion. As a result, the model becomes biased towards positive predictions, preserving recall but impacting precision.

\begin{table}[H]
\caption{Comprehensive performance metrics for ResNet-18 across both training and validation phases. The percentages represent the imbalance levels applied during training, while the validation default distribution remained unchanged.}
\label{tab:resnet18_performance_metrics}
\vskip 0.15in
\begin{center}
\begin{small}
\begin{sc}
\resizebox{\columnwidth}{!}{%
\begin{tabular}{llrrrrrrrrrr}
\toprule
     & & \multicolumn{5}{c}{\textbf{Train}} & \multicolumn{5}{c}{\textbf{Val}} \\
     & & \textbf{1\%} & \textbf{2\%} & \textbf{5\%} & \textbf{15\%} & \textbf{45\%} & \textbf{1\%} & \textbf{2\%} & \textbf{5\%} & \textbf{15\%} & \textbf{45\%}\\
\midrule
\multirow{6}{*}{Precision$\uparrow$} & \textit{Supervised}  & \cellcolor{gray!20}0.055 & \cellcolor{gray!20}0.114 & \cellcolor{gray!20}0.282 & \cellcolor{gray!20}0.614 & \cellcolor{gray!20}0.918 & \cellcolor{gray!20}0.834 & \cellcolor{gray!20}0.840 & \cellcolor{gray!20}0.861 & \cellcolor{gray!20}0.877 & \cellcolor{gray!20}0.893 \\
\cmidrule(lr){2-12}
     & \textit{BYOL}  & 0.027 & 0.053 & 0.143 & 0.372 & 0.782 & 0.715 & 0.715 & 0.747 & 0.750 & 0.801 \\
     & \textit{Barlow Twins}  & 0.032 & 0.067 & 0.167 & 0.410 & 0.805 & 0.749 & 0.769 & 0.764 & 0.772 & 0.811 \\
     & \textit{DINO}  & 0.024 & 0.043 & 0.107 & 0.309 & 0.719 & 0.571 & 0.422 & 0.664 & 0.693 & 0.738 \\
     & \textit{SimCLR}  & 0.028 & 0.066 & 0.168 & 0.433 & 0.808 & 0.715 & 0.762 & 0.770 & 0.786 & 0.817 \\
\midrule
\multirow{6}{*}{Recall$\uparrow$} & \textit{Supervised}  & \cellcolor{gray!20}0.893 & \cellcolor{gray!20}0.895 & \cellcolor{gray!20}0.910 & \cellcolor{gray!20}0.928 & \cellcolor{gray!20}0.927 & \cellcolor{gray!20}0.827 & \cellcolor{gray!20}0.854 & \cellcolor{gray!20}0.869 & \cellcolor{gray!20}0.887 & \cellcolor{gray!20}0.903 \\
\cmidrule(lr){2-12}
     & \textit{BYOL}  & 0.805 & 0.822 & 0.807 & 0.811 & 0.777 & 0.820 & 0.829 & 0.802 & 0.820 & 0.784 \\
     & \textit{Barlow Twins}  & 0.771 & 0.782 & 0.791 & 0.802 & 0.777 & 0.787 & 0.781 & 0.815 & 0.821 & 0.796 \\
     & \textit{DINO}  & 0.663 & 0.670 & 0.669 & 0.733 & 0.711 & 0.026 & 0.017 & 0.683 & 0.747 & 0.716 \\
     & \textit{SimCLR}  & 0.748 & 0.794 & 0.808 & 0.808 & 0.793 & 0.766 & 0.795 & 0.824 & 0.829 & 0.812 \\
\midrule
\multirow{6}{*}{F1$\uparrow$} & \textit{Supervised}  & \cellcolor{gray!20}0.104 & \cellcolor{gray!20}0.202 & \cellcolor{gray!20}0.431 & \cellcolor{gray!20}0.739 & \cellcolor{gray!20}0.922 & \cellcolor{gray!20}0.831 & \cellcolor{gray!20}0.847 & \cellcolor{gray!20}0.865 & \cellcolor{gray!20}0.882 & \cellcolor{gray!20}0.898 \\
\cmidrule(lr){2-12}
     & \textit{BYOL}  & 0.052 & 0.099 & 0.243 & 0.510 & 0.779 & 0.764 & 0.768 & 0.774 & 0.783 & 0.792 \\
     & \textit{Barlow Twins}  & 0.061 & 0.123 & 0.276 & 0.543 & 0.791 & 0.768 & 0.775 & 0.789 & 0.796 & 0.803 \\
     & \textit{DINO}  & 0.046 & 0.080 & 0.185 & 0.434 & 0.715 & 0.049 & 0.032 & 0.673 & 0.719 & 0.727 \\
     & \textit{SimCLR}  & 0.054 & 0.123 & 0.279 & 0.564 & 0.800 & 0.740 & 0.778 & 0.796 & 0.807 & 0.814 \\
\midrule
\multirow{6}{*}{F2$_{CIW}\uparrow$} & \textit{Supervised}  & \cellcolor{gray!20}0.868 & \cellcolor{gray!20}0.877 & \cellcolor{gray!20}0.886 & \cellcolor{gray!20}0.904 & \cellcolor{gray!20}0.900 & \cellcolor{gray!20}0.838 & \cellcolor{gray!20}0.849 & \cellcolor{gray!20}0.880 & \cellcolor{gray!20}0.904 & \cellcolor{gray!20}0.879 \\
\cmidrule(lr){2-12}
     & \textit{BYOL}  & 0.836 & 0.841 & 0.825 & 0.831 & 0.786 & 0.837 & 0.844 & 0.819 & 0.825 & 0.785 \\
     & \textit{Barlow Twins}  & 0.828 & 0.822 & 0.828 & 0.825 & 0.788 & 0.817 & 0.805 & 0.817 & 0.822 & 0.780 \\
     & \textit{DINO}  & 0.757 & 0.719 & 0.733 & 0.761 & 0.699 & 0.566 & 0.402 & 0.712 & 0.777 & 0.752 \\
     & \textit{SimCLR}  & 0.792 & 0.821 & 0.834 & 0.830 & 0.807 & 0.775 & 0.813 & 0.826 & 0.822 & 0.813 \\
\midrule
\multirow{6}{*}{F1$_{Normal}\uparrow$} & \textit{Supervised}  & \cellcolor{gray!20}0.916 & \cellcolor{gray!20}0.922 & \cellcolor{gray!20}0.933 & \cellcolor{gray!20}0.939 & \cellcolor{gray!20}0.936 & \cellcolor{gray!20}0.850 & \cellcolor{gray!20}0.861 & \cellcolor{gray!20}0.879 & \cellcolor{gray!20}0.894 & \cellcolor{gray!20}0.908 \\
\cmidrule(lr){2-12}
     & \textit{BYOL}  & 0.826 & 0.821 & 0.849 & 0.846 & 0.821 & 0.758 & 0.759 & 0.783 & 0.789 & 0.818 \\
     & \textit{Barlow Twins}  & 0.865 & 0.873 & 0.879 & 0.870 & 0.834 & 0.782 & 0.796 & 0.799 & 0.807 & 0.828 \\
     & \textit{DINO}  & 0.832 & 0.812 & 0.820 & 0.808 & 0.769 & 0.689 & 0.686 & 0.700 & 0.730 & 0.763 \\
     & \textit{SimCLR}  & 0.849 & 0.869 & 0.878 & 0.880 & 0.839 & 0.752 & 0.793 & 0.806 & 0.818 & 0.836 \\
\bottomrule
\end{tabular}%
}
\end{sc}
\end{small}
\end{center}
\vskip -0.1in
\end{table}

\clearpage

\begin{table}[H]
\caption{Comprehensive performance metrics for ViT-Tiny across both training and validation phases. The percentages represent the imbalance levels applied during training, while the validation data proportion remained unchanged.}
\label{tab:vit_tiny_performance_metrics}
\vskip 0.15in
\begin{center}
\begin{small}
\begin{sc}
\resizebox{\columnwidth}{!}{%
\begin{tabular}{llrrrrrrrrrr}
\toprule
     & & \multicolumn{5}{c}{\textbf{Train}} & \multicolumn{5}{c}{\textbf{Val}} \\
     & & \textbf{1\%} & \textbf{2\%} & \textbf{5\%} & \textbf{15\%} & \textbf{45\%} & \textbf{1\%} & \textbf{2\%} & \textbf{5\%} & \textbf{15\%} & \textbf{45\%}\\
\midrule
\multirow{6}{*}{Precision$\uparrow$} & \textit{Supervised}  & \cellcolor{gray!20}0.030 & \cellcolor{gray!20}0.065 & \cellcolor{gray!20}0.209 & \cellcolor{gray!20}0.520 & \cellcolor{gray!20}0.879 & \cellcolor{gray!20}0.735 & \cellcolor{gray!20}0.757 & \cellcolor{gray!20}0.816 & \cellcolor{gray!20}0.842 & \cellcolor{gray!20}0.873 \\
\cmidrule(lr){2-12}
     & \textit{BYOL}  & 0.024 & 0.055 & 0.152 & 0.356 & 0.755 & 0.687 & 0.723 & 0.751 & 0.737 & 0.770 \\
     & \textit{Barlow Twins}  & 0.024 & 0.058 & 0.136 & 0.375 & 0.780 & 0.681 & 0.729 & 0.733 & 0.753 & 0.792 \\
     & \textit{DINO}  & 0.018 & 0.041 & 0.052 & 0.307 & 0.714 & 0.625 & 0.651 & 0.474 & 0.692 & 0.734 \\
     & \textit{MAE}  & 0.010 & 0.022 & 0.063 & 0.154 & 0.223 & 0.469 & 0.503 & 0.530 & 0.479 & 0.232 \\
     & \textit{SimCLR}  & 0.033 & 0.063 & 0.155 & 0.373 & 0.776 & 0.752 & 0.746 & 0.753 & 0.747 & 0.788 \\
\midrule
\multirow{6}{*}{Recall$\uparrow$} & \textit{Supervised}  & \cellcolor{gray!20}0.807 & \cellcolor{gray!20}0.808 & \cellcolor{gray!20}0.855 & \cellcolor{gray!20}0.880 & \cellcolor{gray!20}0.890 & \cellcolor{gray!20}0.804 & \cellcolor{gray!20}0.816 & \cellcolor{gray!20}0.847 & \cellcolor{gray!20}0.875 & \cellcolor{gray!20}0.882 \\
\cmidrule(lr){2-12}
     & \textit{BYOL}  & 0.657 & 0.746 & 0.749 & 0.770 & 0.690 & 0.666 & 0.753 & 0.752 & 0.777 & 0.694 \\
     & \textit{Barlow Twins}  & 0.677 & 0.731 & 0.754 & 0.779 & 0.744 & 0.682 & 0.748 & 0.762 & 0.787 & 0.752 \\
     & \textit{DINO}  & 0.679 & 0.710 & 0.674 & 0.715 & 0.695 & 0.773 & 0.719 & 0.662 & 0.717 & 0.703 \\
     & \textit{MAE}  & 0.496 & 0.500 & 0.588 & 0.705 & 0.013 & 0.567 & 0.864 & 0.617 & 0.719 & 0.013 \\
     & \textit{SimCLR}  & 0.761 & 0.783 & 0.788 & 0.805 & 0.776 & 0.772 & 0.792 & 0.790 & 0.812 & 0.784 \\
\midrule
\multirow{6}{*}{F1$\uparrow$} & \textit{Supervised}  & \cellcolor{gray!20}0.059 & \cellcolor{gray!20}0.120 & \cellcolor{gray!20}0.335 & \cellcolor{gray!20}0.654 & \cellcolor{gray!20}0.884 & \cellcolor{gray!20}0.768 & \cellcolor{gray!20}0.785 & \cellcolor{gray!20}0.831 & \cellcolor{gray!20}0.858 & \cellcolor{gray!20}0.878 \\
\cmidrule(lr){2-12}
     & \textit{BYOL}  & 0.046 & 0.102 & 0.252 & 0.487 & 0.721 & 0.676 & 0.738 & 0.752 & 0.757 & 0.730 \\
     & \textit{Barlow Twins}  & 0.046 & 0.107 & 0.231 & 0.506 & 0.762 & 0.682 & 0.738 & 0.747 & 0.769 & 0.772 \\
     & \textit{DINO}  & 0.035 & 0.077 & 0.096 & 0.430 & 0.704 & 0.691 & 0.683 & 0.553 & 0.704 & 0.719 \\
     & \textit{MAE}  & 0.020 & 0.043 & 0.114 & 0.253 & 0.025 & 0.513 & 0.636 & 0.570 & 0.575 & 0.024 \\
     & \textit{SimCLR}  & 0.062 & 0.116 & 0.259 & 0.509 & 0.776 & 0.762 & 0.768 & 0.771 & 0.778 & 0.786 \\
\midrule
\multirow{6}{*}{F2$_{CIW}\uparrow$} & \textit{Supervised}  & \cellcolor{gray!20}0.824 & \cellcolor{gray!20}0.825 & \cellcolor{gray!20}0.848 & \cellcolor{gray!20}0.869 & \cellcolor{gray!20}0.868 & \cellcolor{gray!20}0.837 & \cellcolor{gray!20}0.812 & \cellcolor{gray!20}0.831 & \cellcolor{gray!20}0.870 & \cellcolor{gray!20}0.881 \\
\cmidrule(lr){2-12}
     & \textit{BYOL}  & 0.711 & 0.773 & 0.776 & 0.789 & 0.726 & 0.680 & 0.768 & 0.760 & 0.781 & 0.725 \\
     & \textit{Barlow Twins}  & 0.708 & 0.759 & 0.787 & 0.804 & 0.760 & 0.704 & 0.749 & 0.786 & 0.784 & 0.758 \\
     & \textit{DINO}  & 0.722 & 0.744 & 0.715 & 0.729 & 0.680 & 0.733 & 0.736 & 0.690 & 0.724 & 0.695 \\
     & \textit{MAE}  & 0.624 & 0.394 & 0.711 & 0.683 & 0.014 & 0.629 & 0.640 & 0.704 & 0.702 & 0.009 \\
     & \textit{SimCLR}  & 0.792 & 0.809 & 0.807 & 0.815 & 0.785 & 0.766 & 0.795 & 0.798 & 0.801 & 0.788 \\
\midrule
\multirow{6}{*}{F1$_{Normal}\uparrow$} & \textit{Supervised}  & \cellcolor{gray!20}0.849 & \cellcolor{gray!20}0.862 & \cellcolor{gray!20}0.903 & \cellcolor{gray!20}0.912 & \cellcolor{gray!20}0.904 & \cellcolor{gray!20}0.774 & \cellcolor{gray!20}0.794 & \cellcolor{gray!20}0.844 & \cellcolor{gray!20}0.869 & \cellcolor{gray!20}0.890 \\
\cmidrule(lr){2-12}
     & \textit{BYOL}  & 0.839 & 0.847 & 0.869 & 0.840 & 0.789 & 0.719 & 0.757 & 0.778 & 0.771 & 0.781 \\
     & \textit{Barlow Twins}  & 0.834 & 0.858 & 0.850 & 0.851 & 0.813 & 0.716 & 0.760 & 0.766 & 0.785 & 0.806 \\
     & \textit{DINO}  & 0.785 & 0.791 & 0.507 & 0.810 & 0.764 & 0.655 & 0.688 & 0.419 & 0.727 & 0.759 \\
     & \textit{MAE}  & 0.615 & 0.709 & 0.692 & 0.461 & 0.695 & 0.471 & 0.350 & 0.553 & 0.389 & 0.677 \\
     & \textit{SimCLR}  & 0.869 & 0.862 & 0.867 & 0.847 & 0.816 & 0.782 & 0.781 & 0.786 & 0.785 & 0.810 \\
\bottomrule
\end{tabular}%
}
\end{sc}
\end{small}
\end{center}
\vskip -0.1in
\end{table}

\section{Self-supervised monitoring metrics}
\label{app:ssl_monitor_metrics}

\begin{figure}[H]
    \centering
    \begin{subfigure}[b]{0.9\textwidth}
        \includegraphics[width=\textwidth]{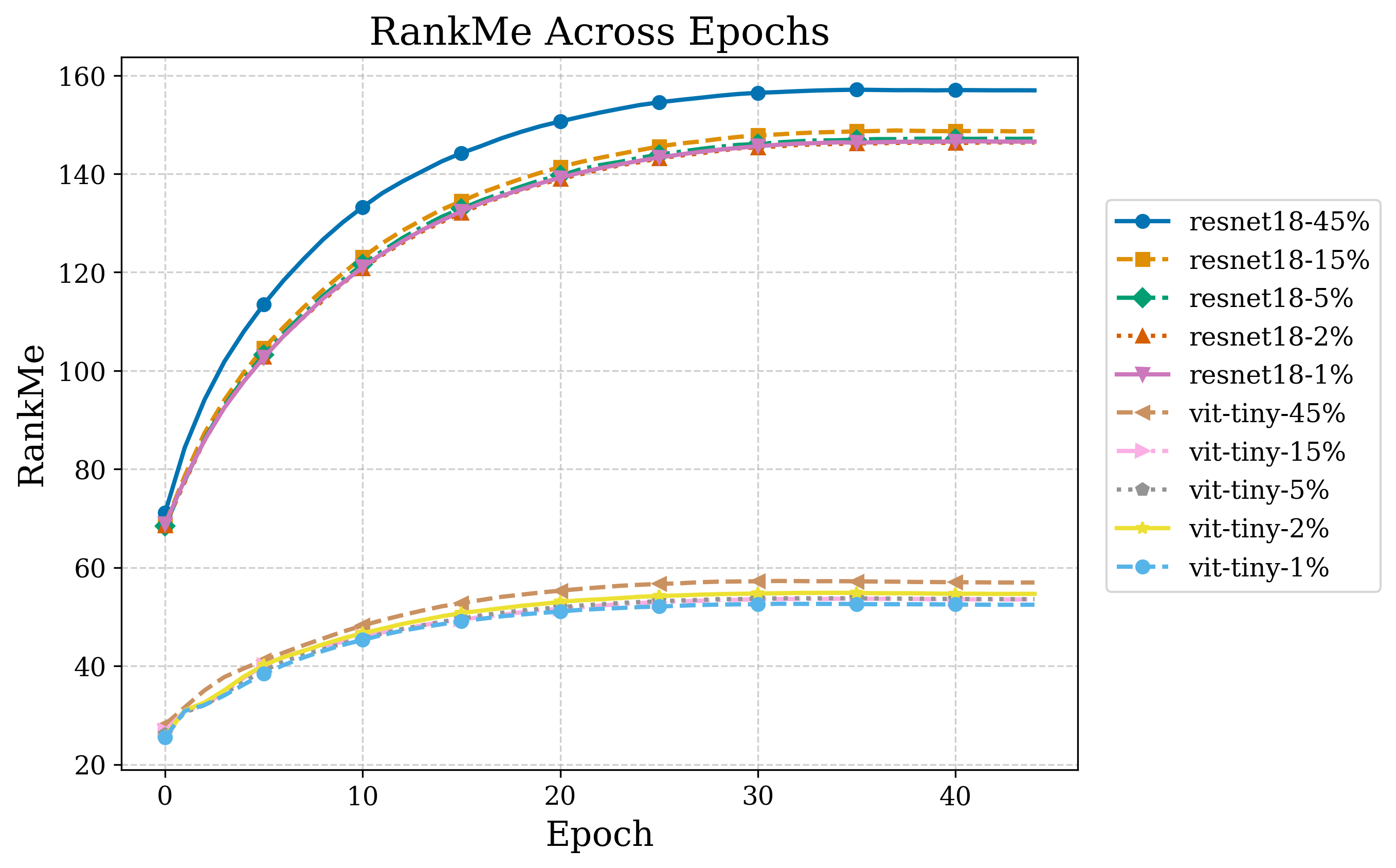}
        \caption{RankMe metric across epochs.}
    \end{subfigure}

    \begin{subfigure}[b]{\textwidth}
        \includegraphics[width=\textwidth]{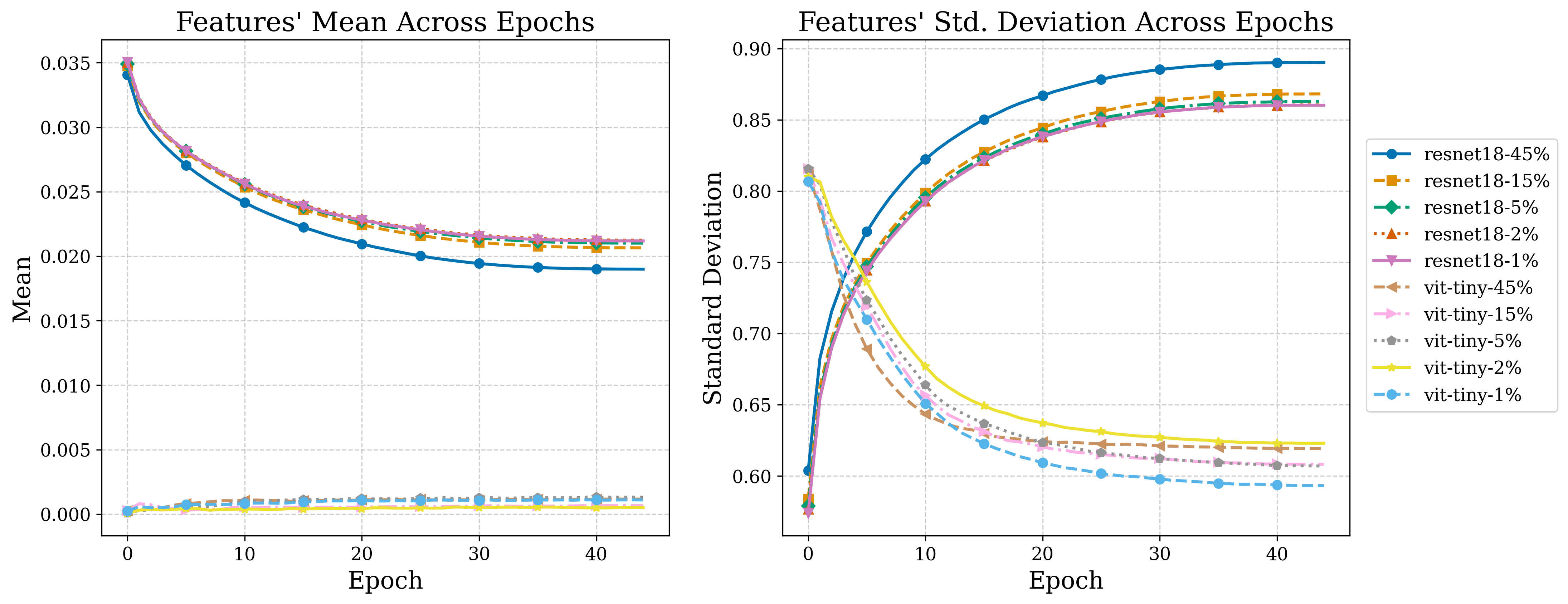}
        \caption{Features' mean and standard deviation across epochs.}
    \end{subfigure}

    \begin{subfigure}[b]{\textwidth}
        \includegraphics[width=\textwidth]{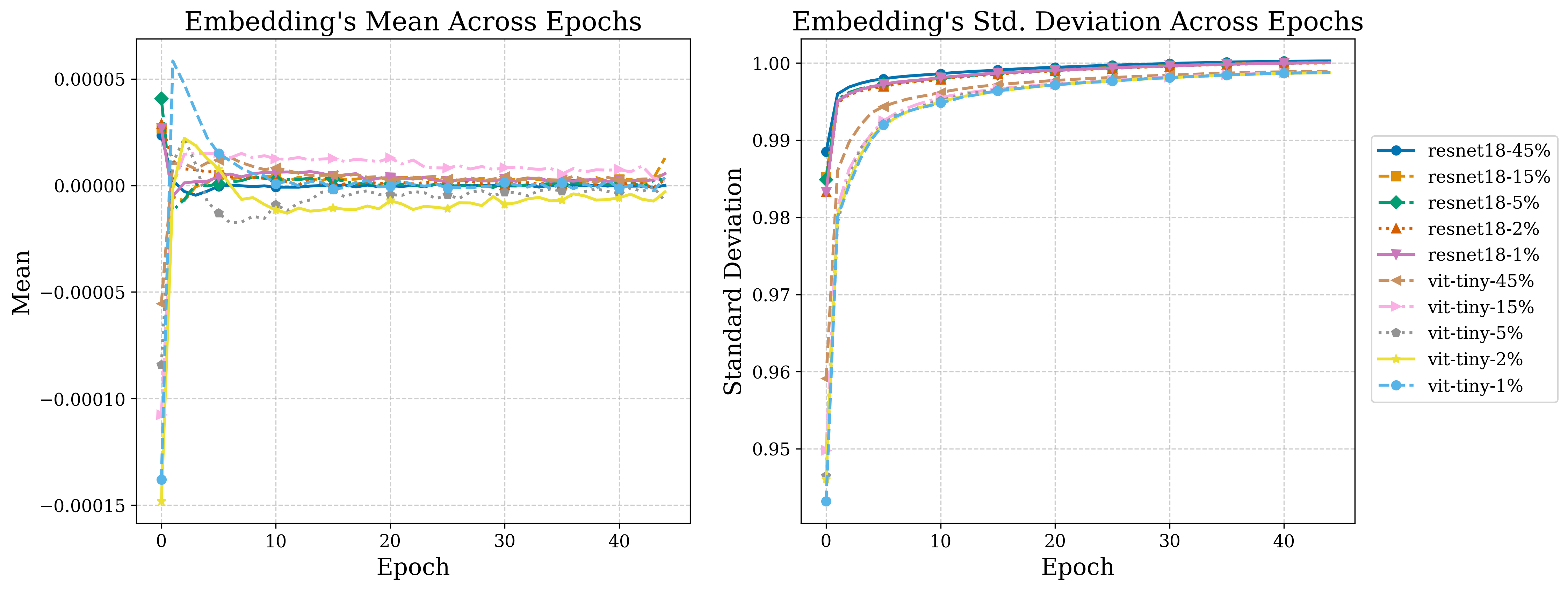}
        \caption{Embedding's mean and standard deviation across epochs.}
    \end{subfigure}

    \caption{Self-supervised monitoring metrics for SimCLR.}
    \label{fig:rankme-simclr}
\end{figure}

\begin{figure}[H]
    \centering
    \begin{subfigure}[b]{0.9\textwidth}
        \includegraphics[width=\textwidth]{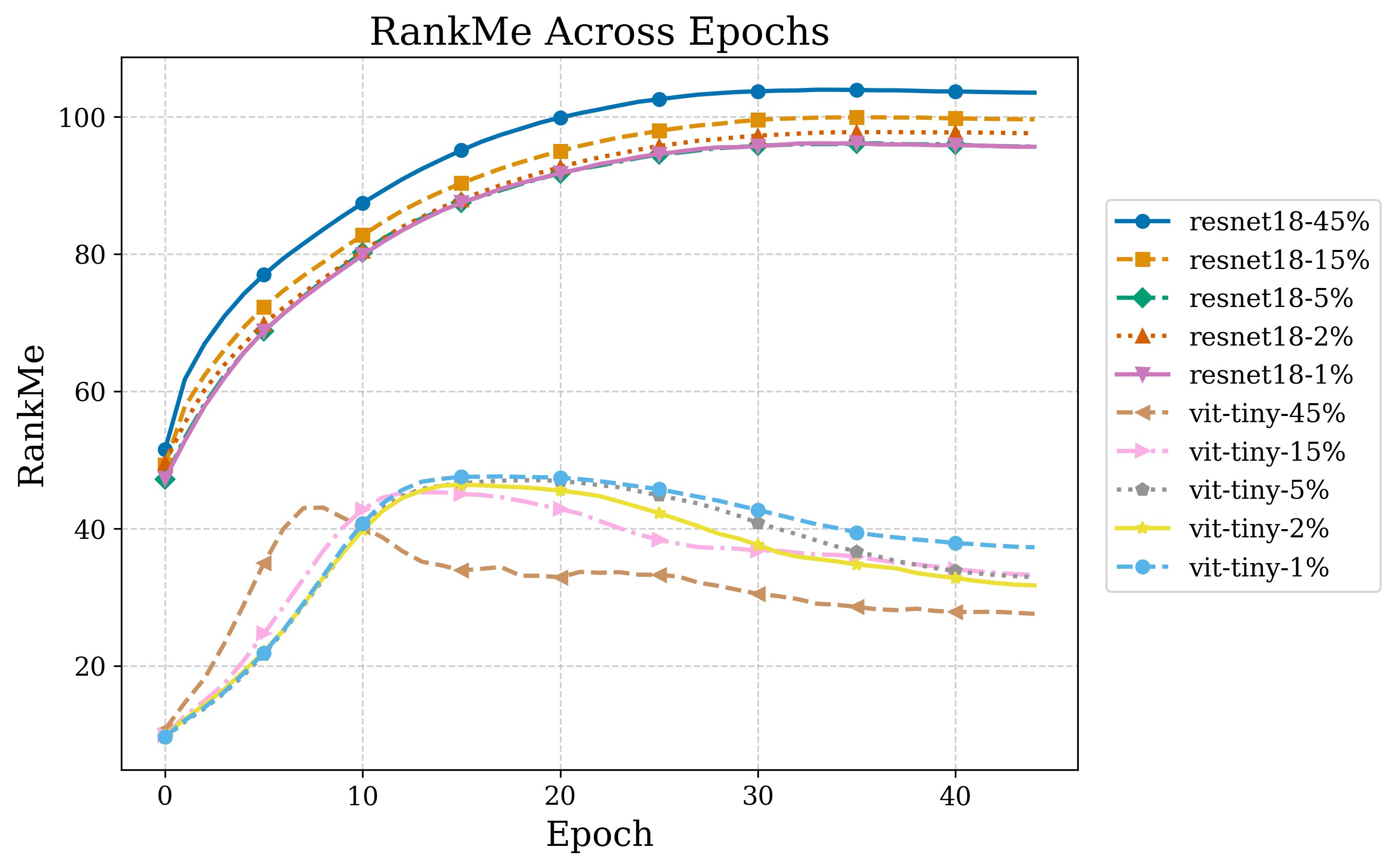}
        \caption{RankMe metric across epochs.}
    \end{subfigure}

    \begin{subfigure}[b]{\textwidth}
        \includegraphics[width=\textwidth]{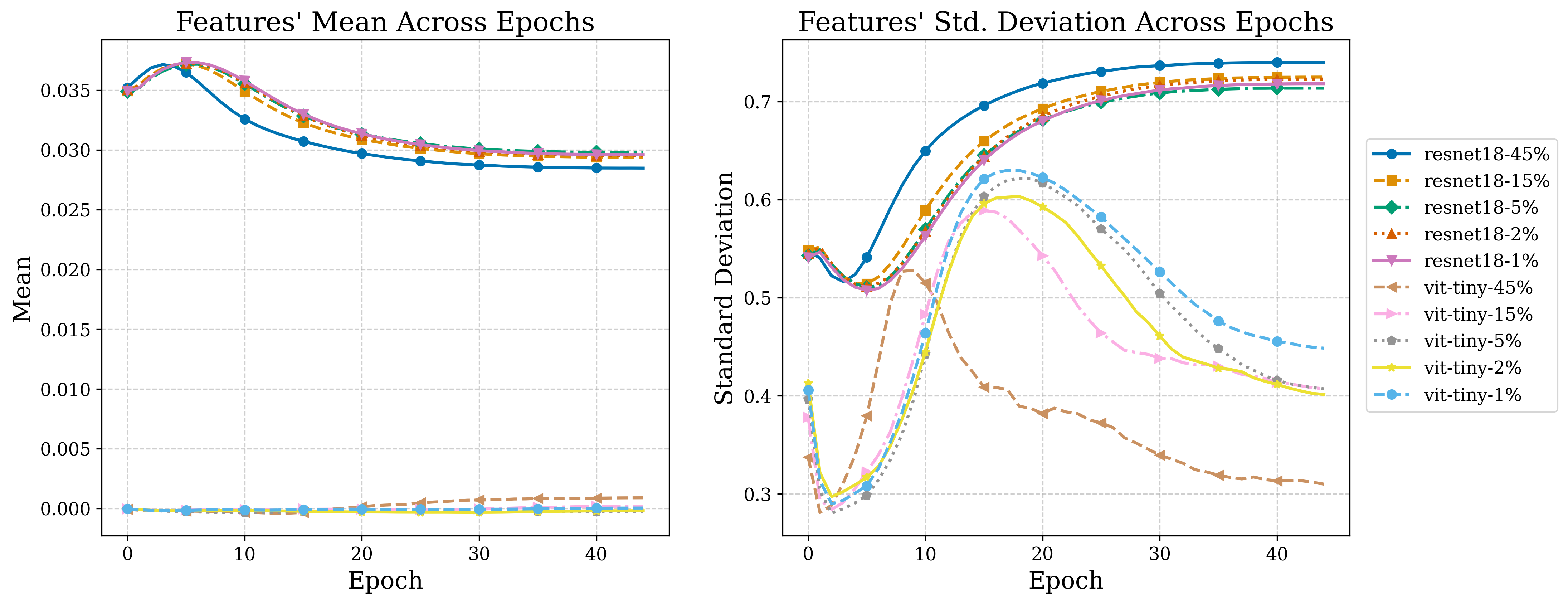}
        \caption{Features' mean and standard deviation across epochs.}
    \end{subfigure}

    \begin{subfigure}[b]{\textwidth}
        \includegraphics[width=\textwidth]{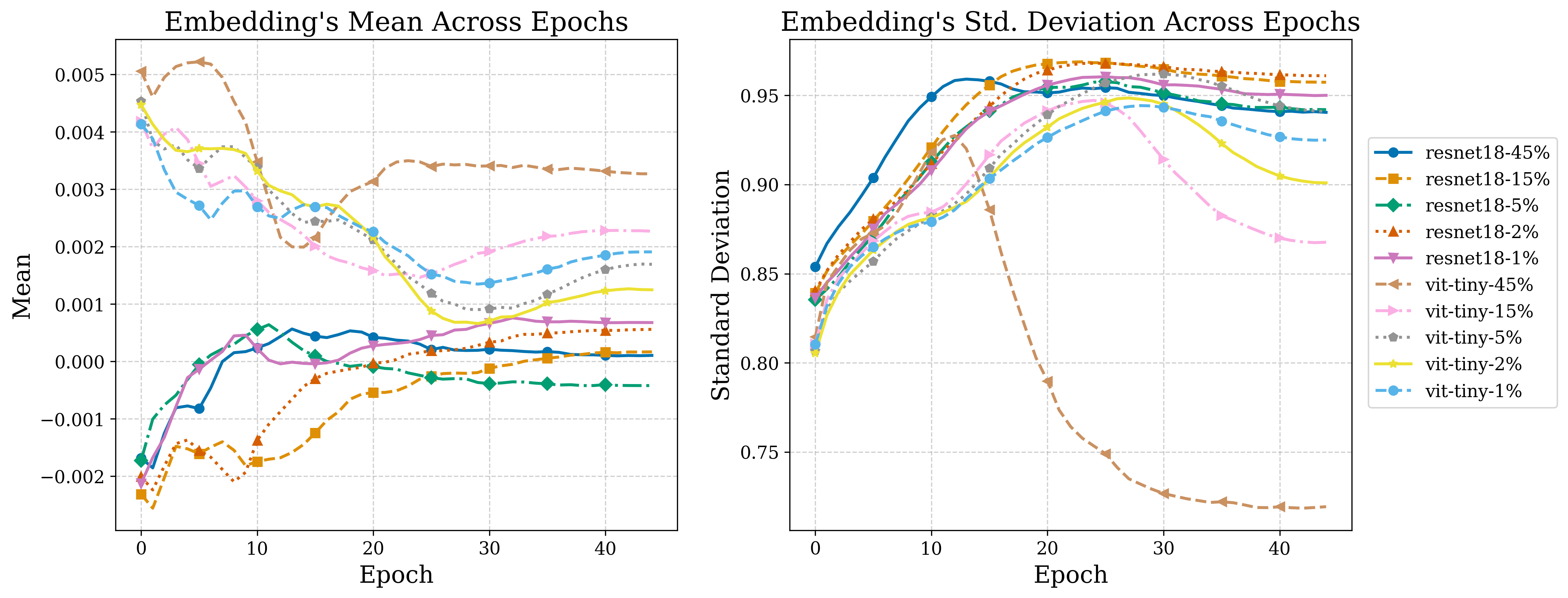}
        \caption{Embedding's mean and standard deviation across epochs.}
    \end{subfigure}

    \caption{Self-supervised monitoring metrics for BYOL.}
    \label{fig:rankme-byol}
\end{figure}

\begin{figure}[H]
    \centering
    \begin{subfigure}[b]{0.9\textwidth}
        \includegraphics[width=\textwidth]{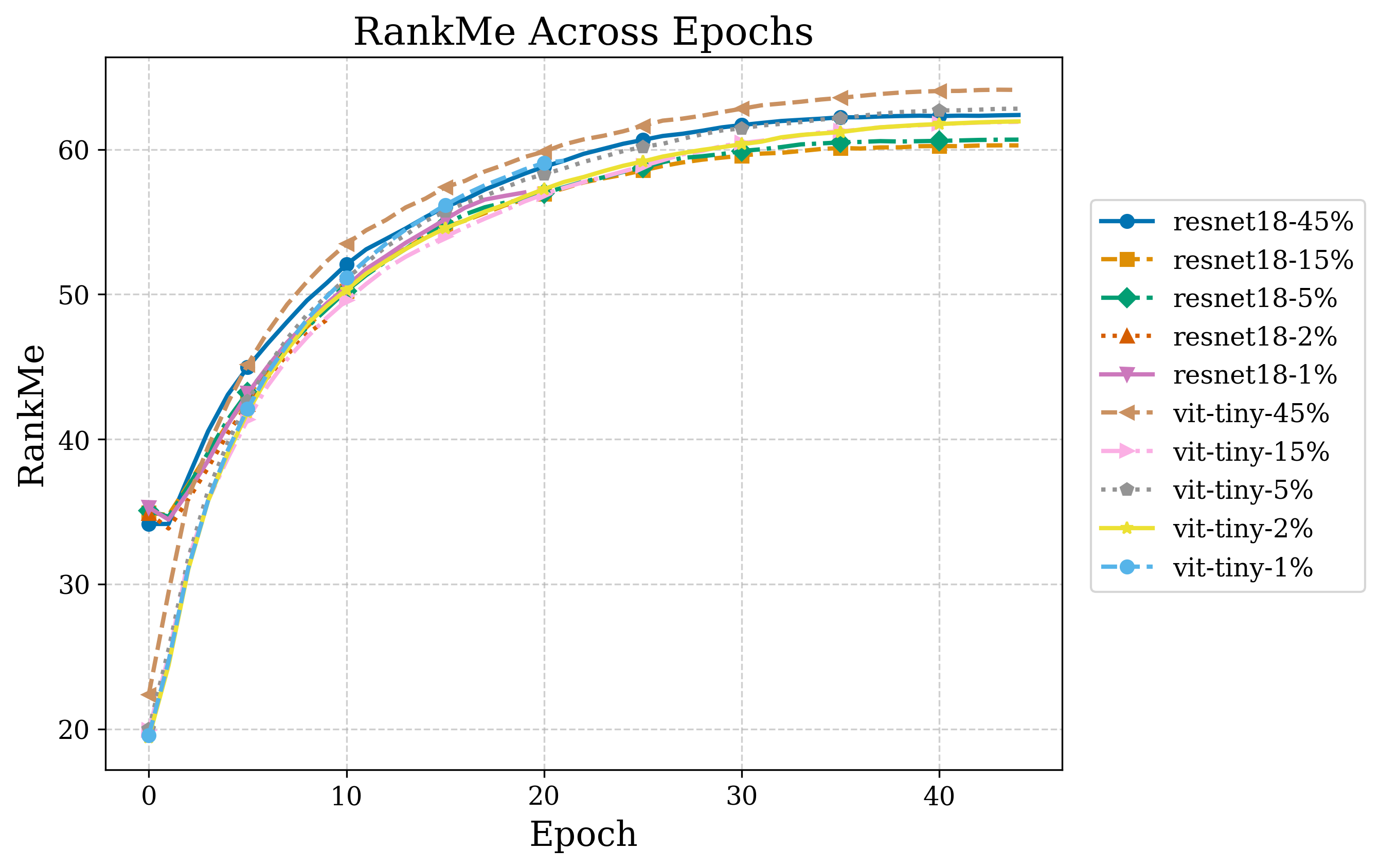}
        \caption{RankMe metric across epochs.}
    \end{subfigure}

    \begin{subfigure}[b]{\textwidth}
        \includegraphics[width=\textwidth]{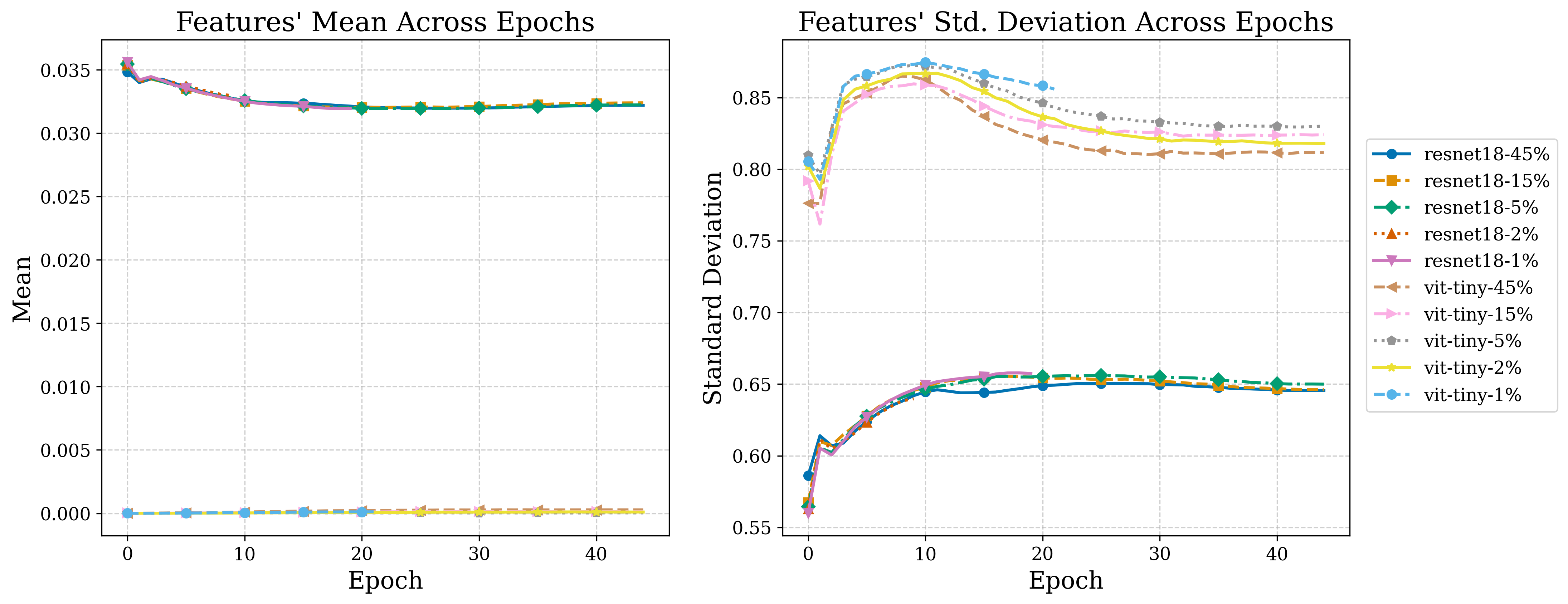}
        \caption{Features' mean and standard deviation across epochs.}
    \end{subfigure}

    \begin{subfigure}[b]{\textwidth}
        \includegraphics[width=\textwidth]{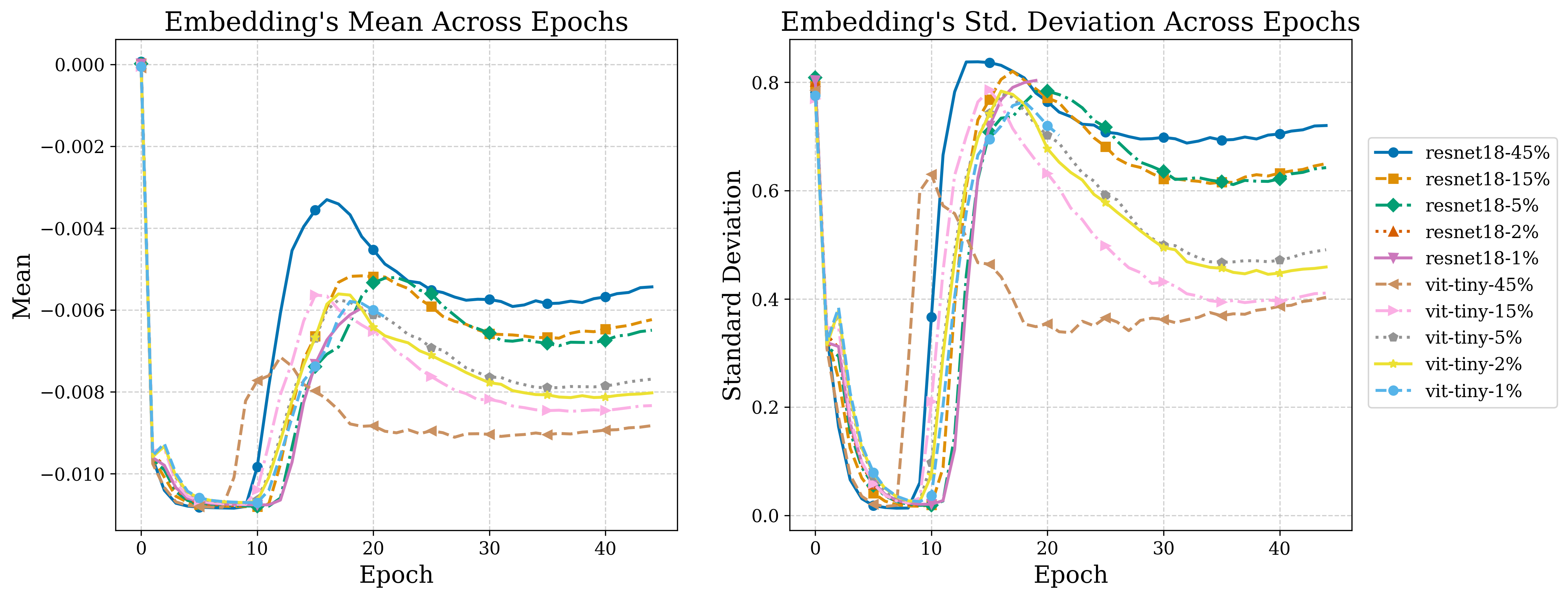}
        \caption{Embedding's mean and standard deviation across epochs.}
    \end{subfigure}

    \caption{Self-supervised monitoring metrics for DINO.}
    \label{fig:rankme-dino}
\end{figure}

When evaluating SimCLR, BYOL, and DINO, significant differences emerge in the behavior of ResNet-18 versus ViT-Tiny architectures. ResNet-18 consistently achieved higher RankMe values and demonstrated steady improvement in feature standard deviation, indicating more stable and robust feature representations. In contrast, ViT-Tiny models showed strong initial performance that often degraded over time, particularly under SimCLR and BYOL, suggesting challenges in maintaining feature consistency (see Figures~\ref{fig:rankme-simclr},~\ref{fig:rankme-byol}). However, under DINO, ViT-Tiny’s performance was more stable, aligning closely with ResNet-18's in terms of RankMe values (see Figure~\ref{fig:rankme-dino}). 

Overall, these findings suggest that ResNet architectures offer superior stability and consistency across different self-supervised learning algorithms, whereas ViT-Tiny models may require tailored optimization techniques to sustain their initial performance levels. This underscores the importance of continuously monitoring self-supervised metrics to ensure robust feature learning and representation stability.

\begin{figure}[H]
    \centering
    \includegraphics[width=\linewidth]{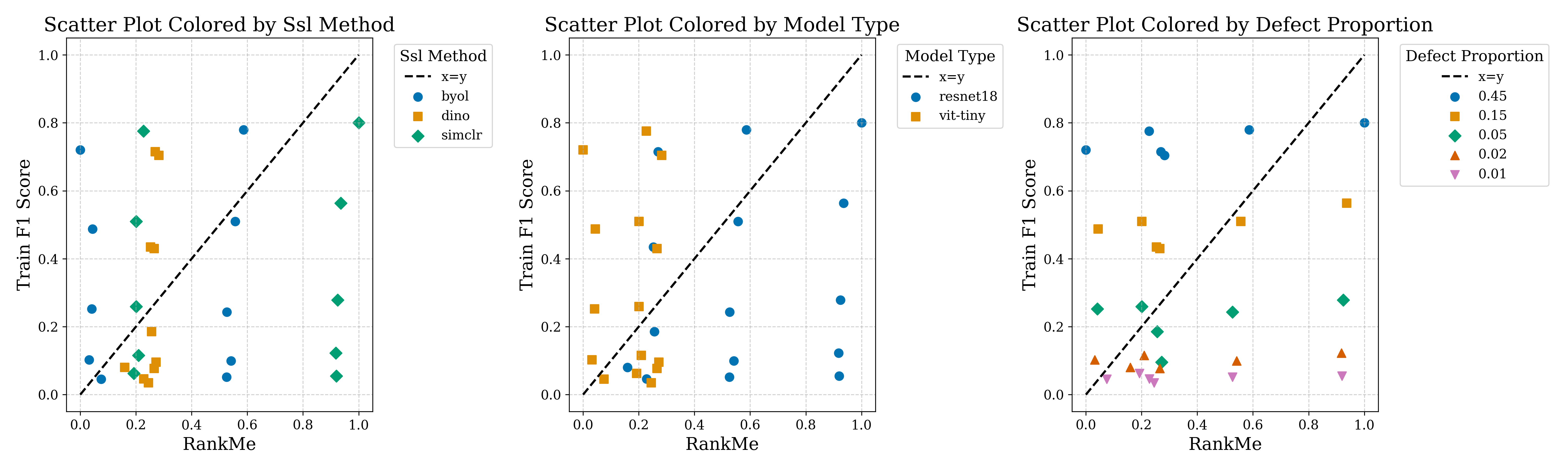}
    \caption{Train F1 scores plotted against re-scaled RankMe metrics. RankMe metrics were re-scaled to enable easier comparisons with F1 scores.}
    \label{fig:train_rankme_scatter}
\end{figure}

\begin{figure}[H]
    \centering
    \includegraphics[width=\linewidth]{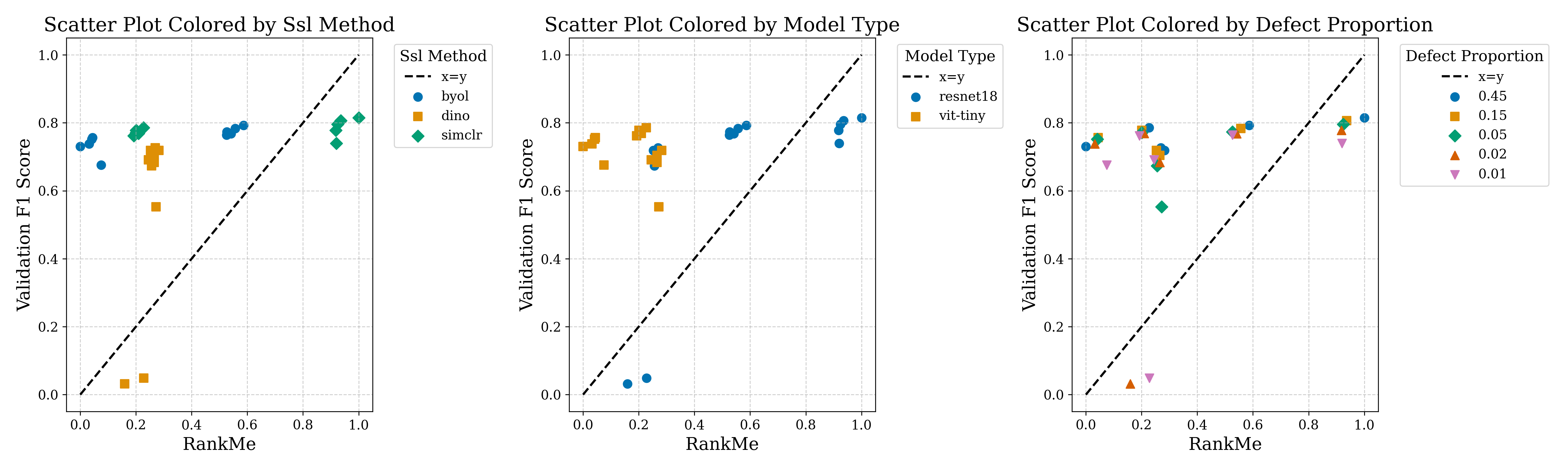}
    \caption{Validation F1 scores plotted against re-scaled RankMe metrics. RankMe metrics were re-scaled to enable easier comparisons with F1 scores.}
    \label{fig:val_rankme_scatter}
\end{figure}

Furthermore, scatter plots comparing RankMe with train and validation F1 scores—grouped by SSL method, model type, and defect proportion—show no correlation (Figures~\ref{fig:train_rankme_scatter},~\ref{fig:val_rankme_scatter}, indicating that RankMe is not a reliable predictor of downstream performance. This challenges RankMe’s intended role and highlights a gap in its effectiveness within the SSL domain. Nonetheless, distinct RankMe differences between ResNet-18 and ViT-Tiny architectures persist, reinforcing the architectural distinctions previously observed.

\newpage

\subsection{SimCLR}

\newpage

\subsection{BYOL}

\newpage

\subsection{DINO}

\newpage

\subsection{RankMe scatter plots}

\newpage

\section{Sewer-ML samples}
\label{app:train-val-image-samples}

\begin{table}[H]
\centering
\resizebox{\textwidth}{!}{
\begin{tabular}{>{\centering\arraybackslash}m{3cm}|>{\centering\arraybackslash}c>{\centering\arraybackslash}c>{\centering\arraybackslash}c>{\centering\arraybackslash}c|>{\centering\arraybackslash}c>{\centering\arraybackslash}c>{\centering\arraybackslash}c>{\centering\arraybackslash}c}
\textbf{Defect Type} & \multicolumn{4}{c}{\textbf{Train}} & \multicolumn{4}{c}{\textbf{Validation}}  \\ 
\cmidrule(lr){1-2}
\cmidrule(lr){2-6}
\cmidrule(lr){6-9}
\textbf{RB} & \includegraphics[width=0.1\textwidth]{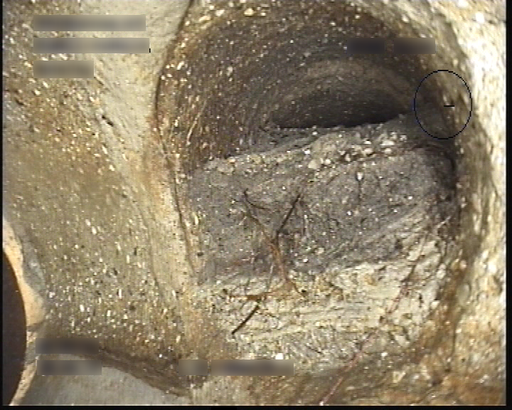} & \includegraphics[width=0.1\textwidth]{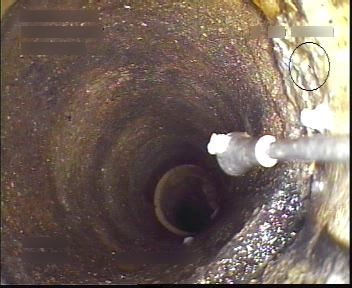} & \includegraphics[width=0.1\textwidth]{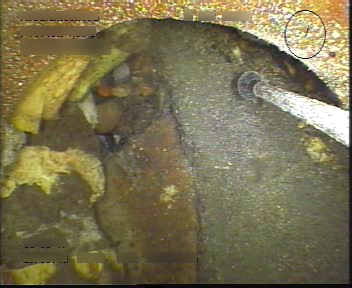} & \includegraphics[width=0.1\textwidth]{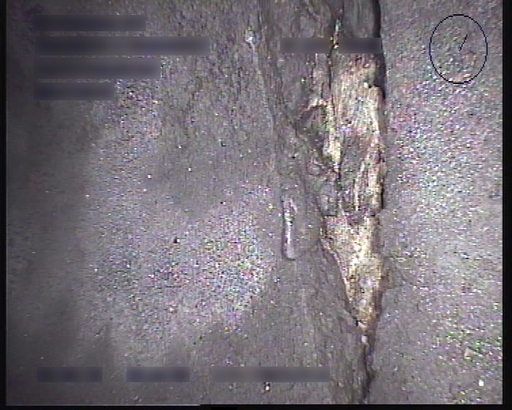} & \includegraphics[width=0.1\textwidth]{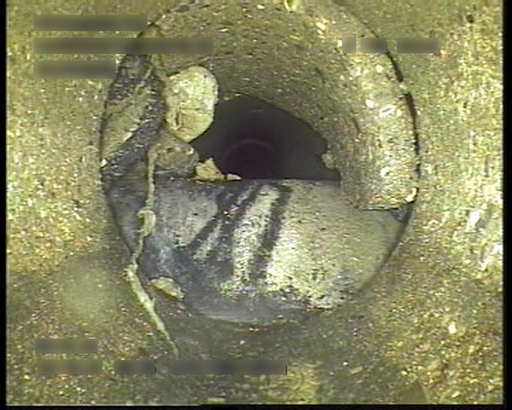} & \includegraphics[width=0.1\textwidth]{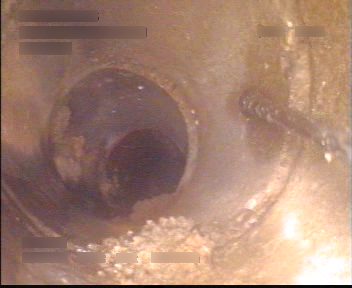} & \includegraphics[width=0.1\textwidth]{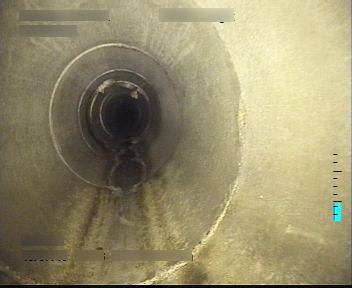} & \includegraphics[width=0.1\textwidth]{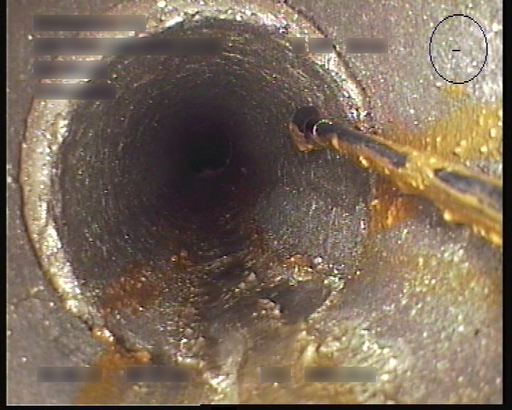} \\
\textbf{OS} & \includegraphics[width=0.1\textwidth]{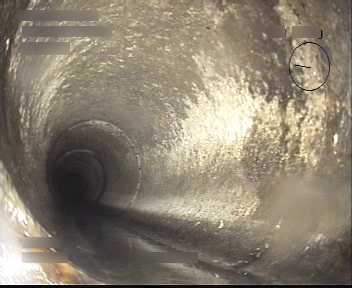} & \includegraphics[width=0.1\textwidth]{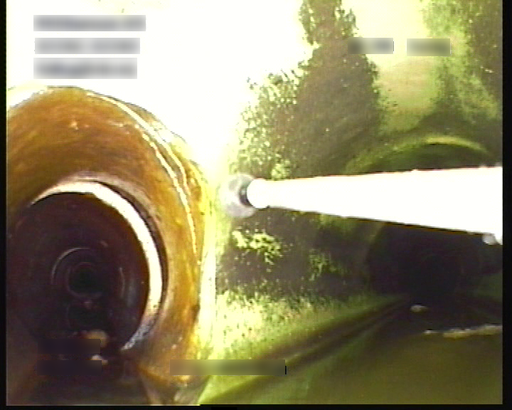} & \includegraphics[width=0.1\textwidth]{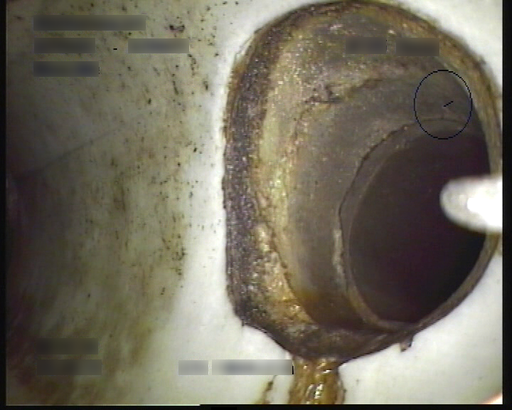} & \includegraphics[width=0.1\textwidth]{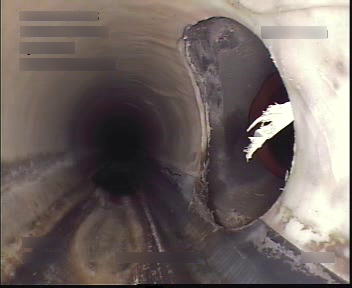} & \includegraphics[width=0.1\textwidth]{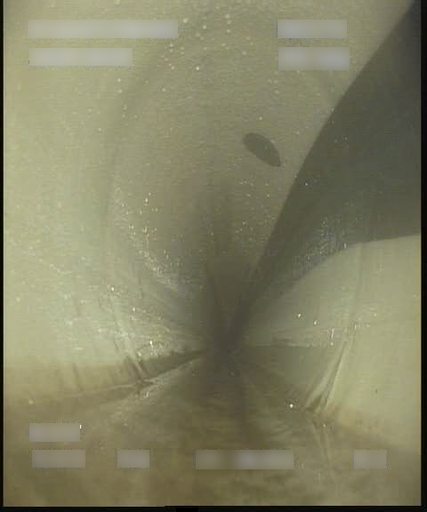} & \includegraphics[width=0.1\textwidth]{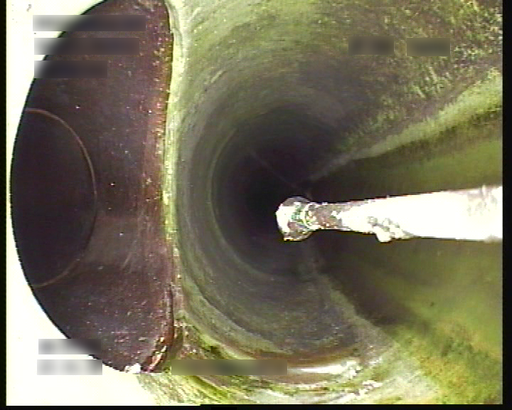} & \includegraphics[width=0.1\textwidth]{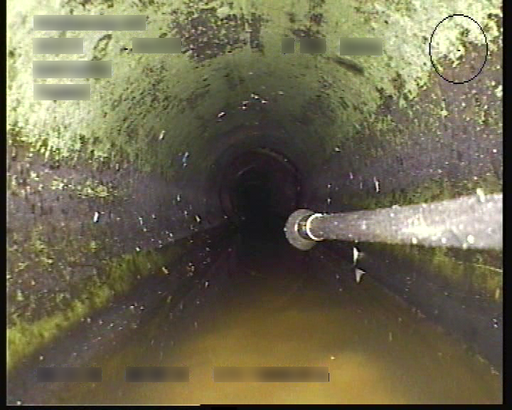} & \includegraphics[width=0.1\textwidth]{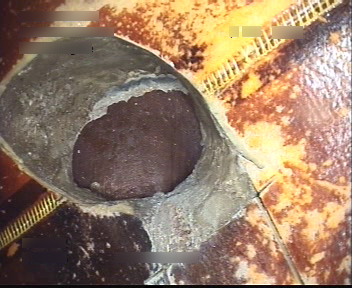} \\
\textbf{FS} & \includegraphics[width=0.1\textwidth]{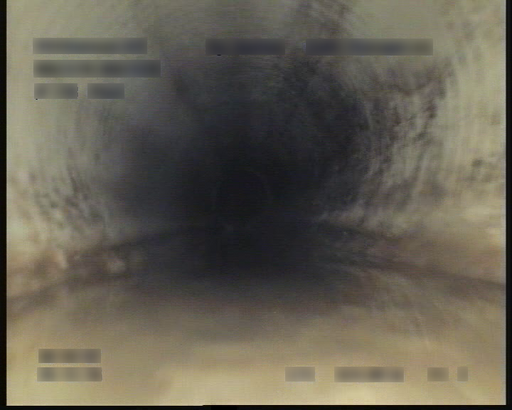} & \includegraphics[width=0.1\textwidth]{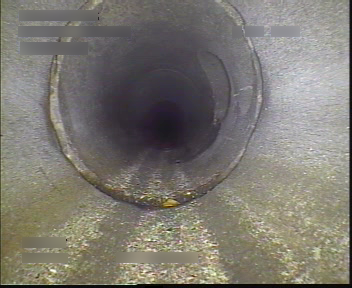} & \includegraphics[width=0.1\textwidth]{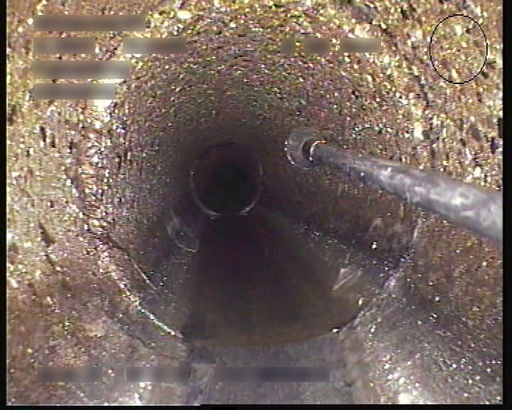} & \includegraphics[width=0.1\textwidth]{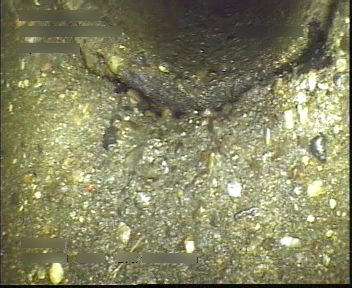} & \includegraphics[width=0.1\textwidth]{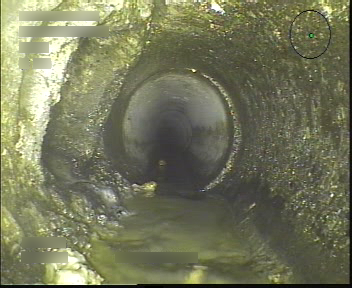} & \includegraphics[width=0.1\textwidth]{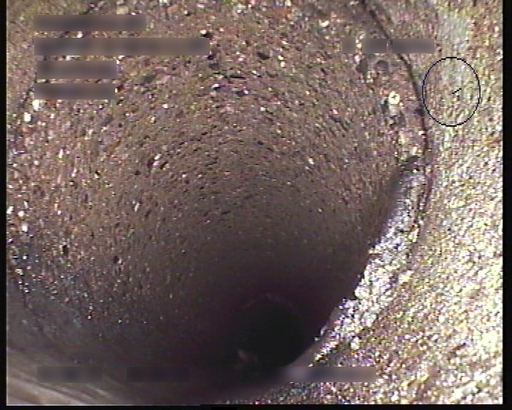} & \includegraphics[width=0.1\textwidth]{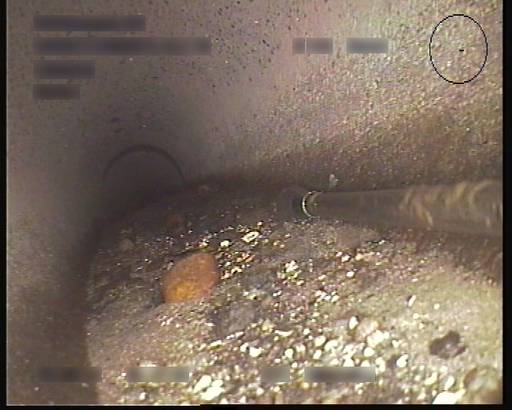} & \includegraphics[width=0.1\textwidth]{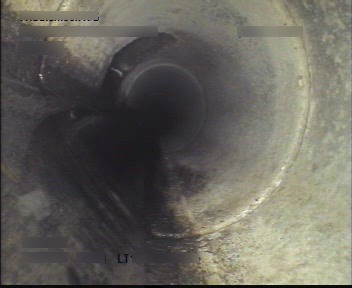} \\
\textbf{OB} & \includegraphics[width=0.1\textwidth]{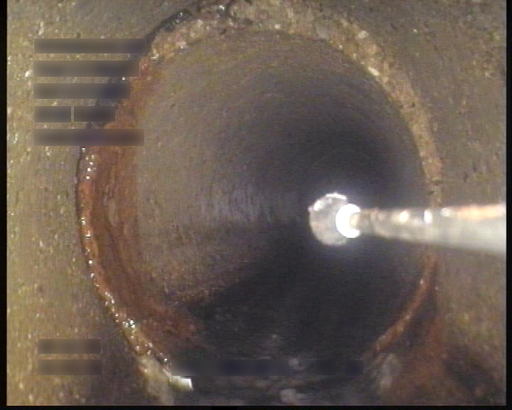} & \includegraphics[width=0.1\textwidth]{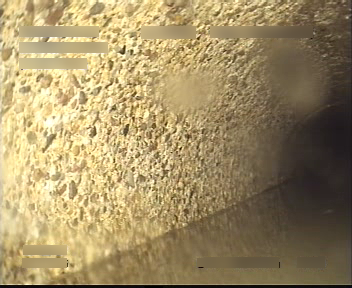} & \includegraphics[width=0.1\textwidth]{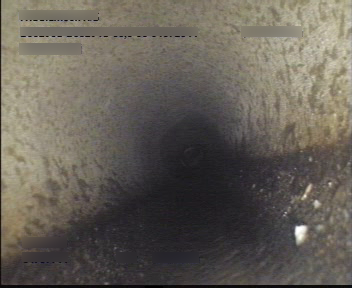} & \includegraphics[width=0.1\textwidth]{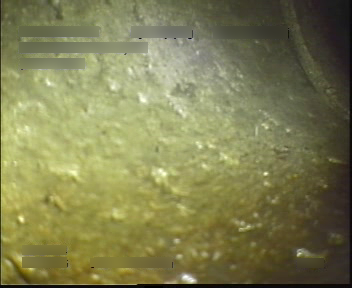} & \includegraphics[width=0.1\textwidth]{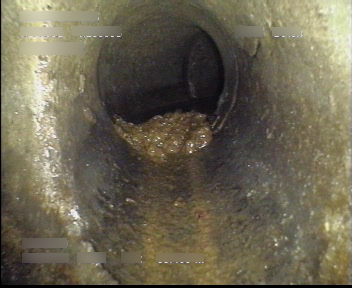} & \includegraphics[width=0.1\textwidth]{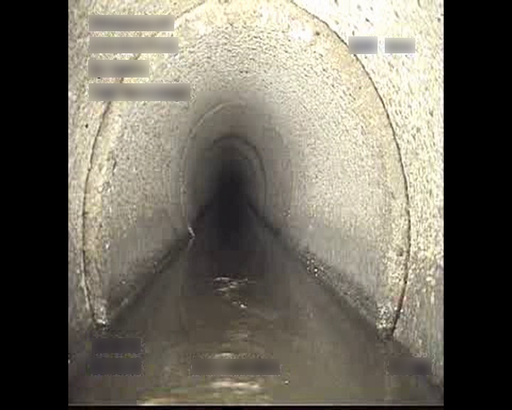} & \includegraphics[width=0.1\textwidth]{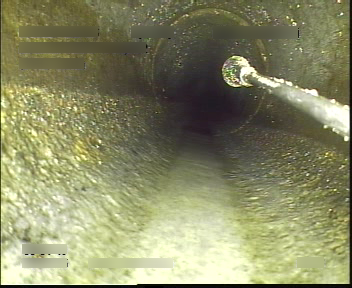} & \includegraphics[width=0.1\textwidth]{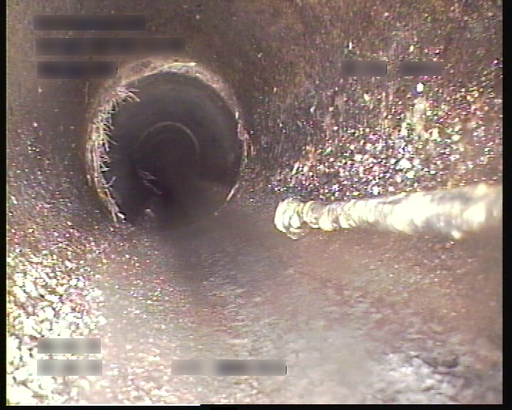} \\
\textbf{OK} & \includegraphics[width=0.1\textwidth]{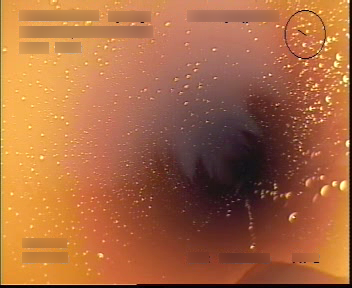} & \includegraphics[width=0.1\textwidth]{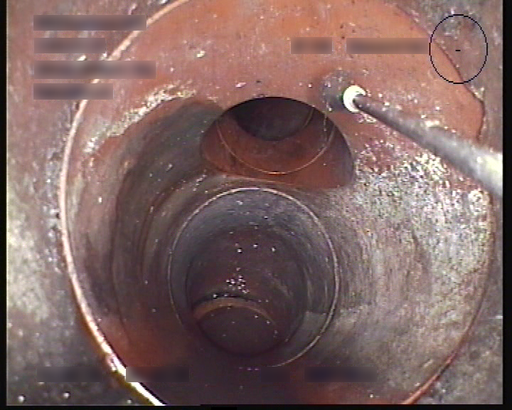} & \includegraphics[width=0.1\textwidth]{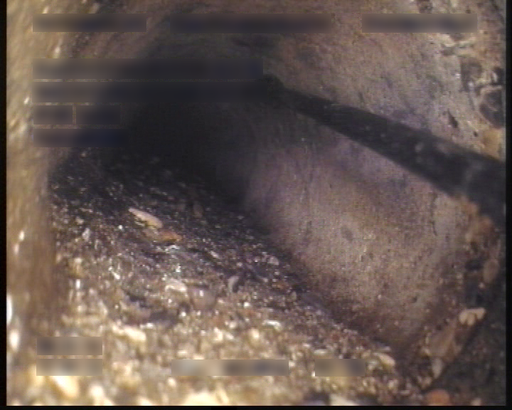} & \includegraphics[width=0.1\textwidth]{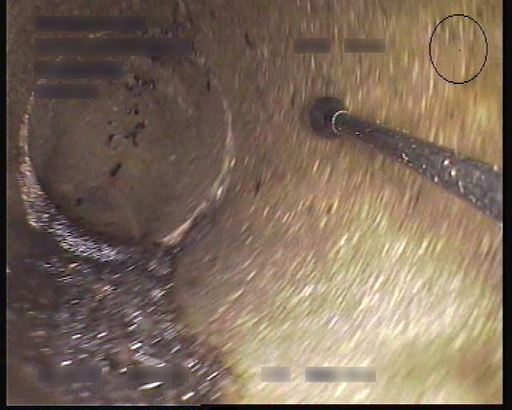} & \includegraphics[width=0.1\textwidth]{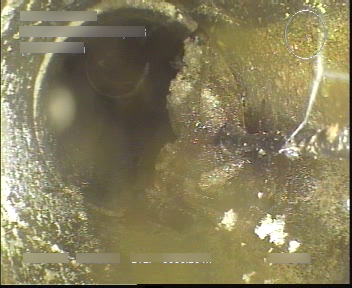} & \includegraphics[width=0.1\textwidth]{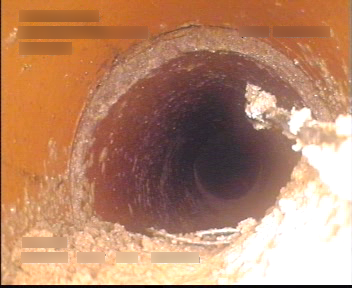} & \includegraphics[width=0.1\textwidth]{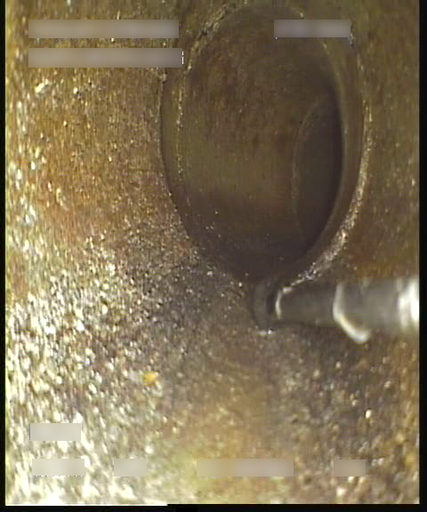} & \includegraphics[width=0.1\textwidth]{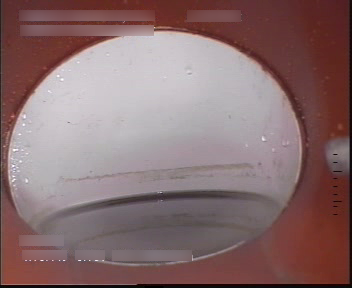} \\
\textbf{PH} & \includegraphics[width=0.1\textwidth]{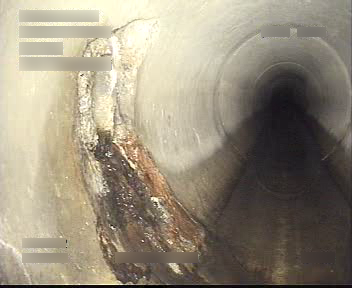} & \includegraphics[width=0.1\textwidth]{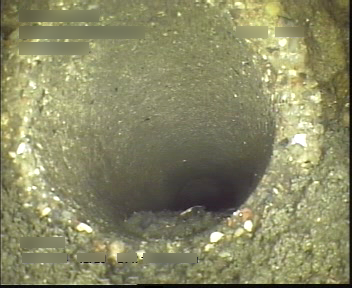} & \includegraphics[width=0.1\textwidth]{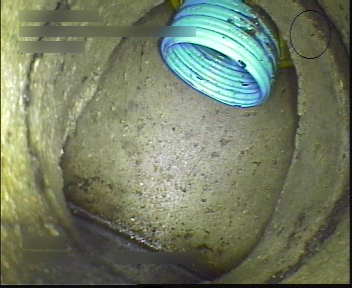} & \includegraphics[width=0.1\textwidth]{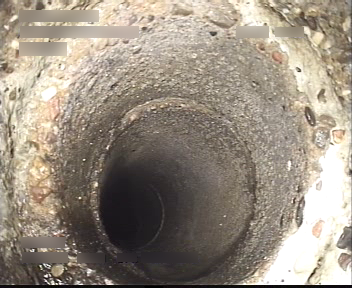} & \includegraphics[width=0.1\textwidth]{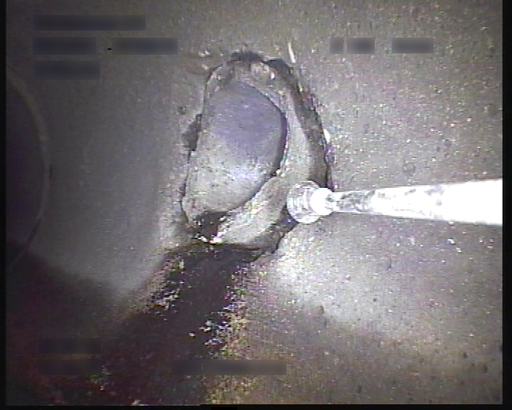} & \includegraphics[width=0.1\textwidth]{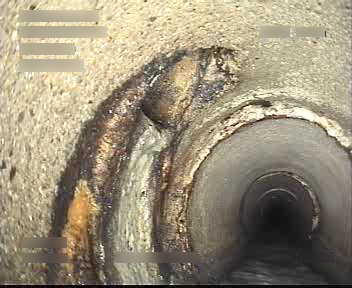} & \includegraphics[width=0.1\textwidth]{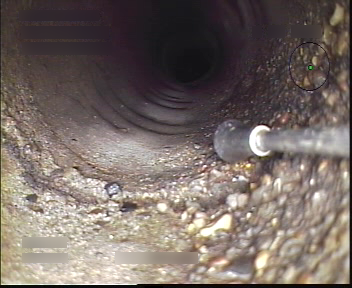} & \includegraphics[width=0.1\textwidth]{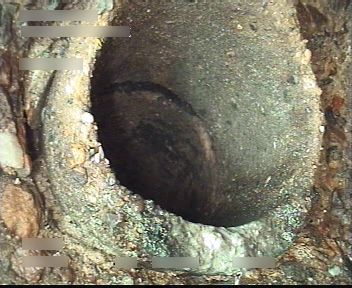} \\
\textbf{PB} & \includegraphics[width=0.1\textwidth]{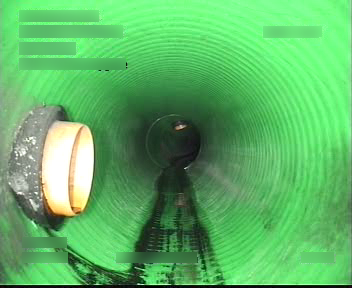} & \includegraphics[width=0.1\textwidth]{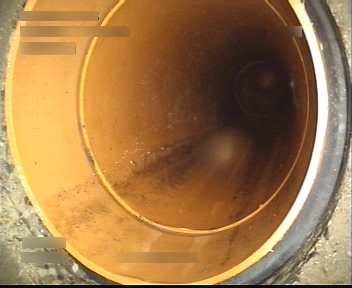} & \includegraphics[width=0.1\textwidth]{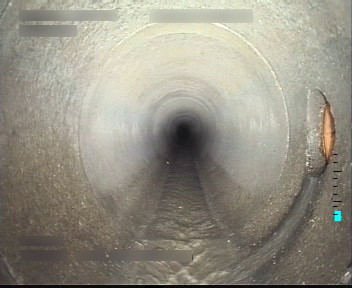} & \includegraphics[width=0.1\textwidth]{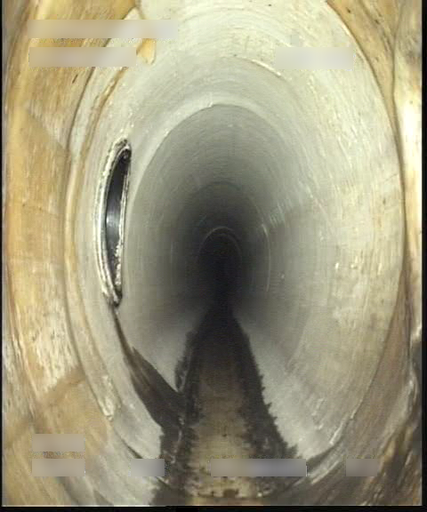} & \includegraphics[width=0.1\textwidth]{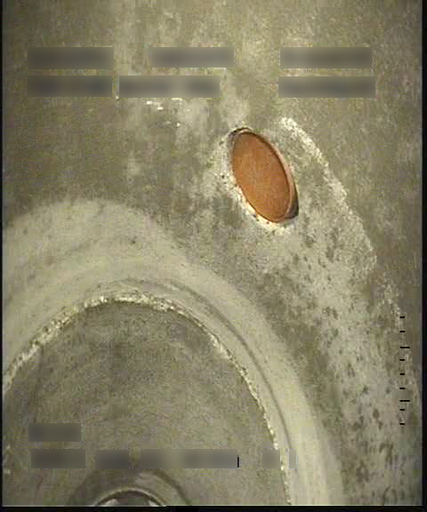} & \includegraphics[width=0.1\textwidth]{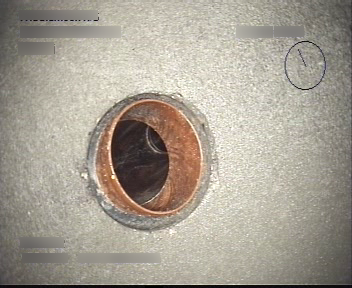} & \includegraphics[width=0.1\textwidth]{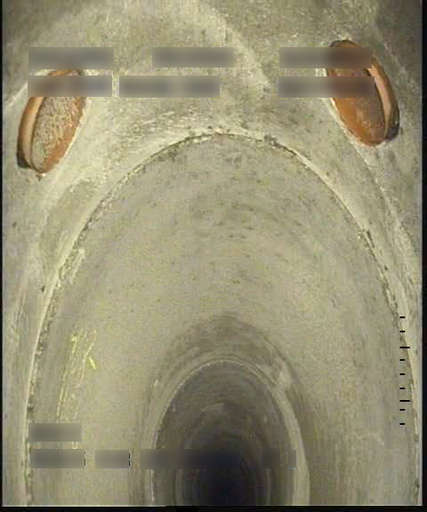} & \includegraphics[width=0.1\textwidth]{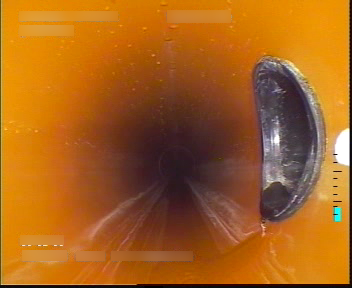} \\
\textbf{OP} & \includegraphics[width=0.1\textwidth]{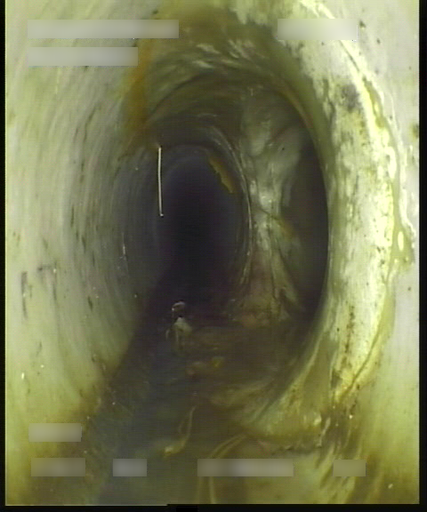} & \includegraphics[width=0.1\textwidth]{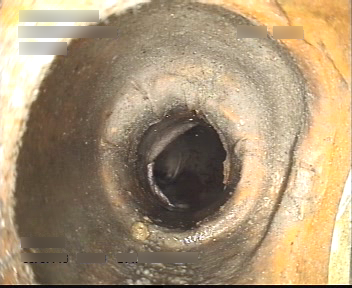} & \includegraphics[width=0.1\textwidth]{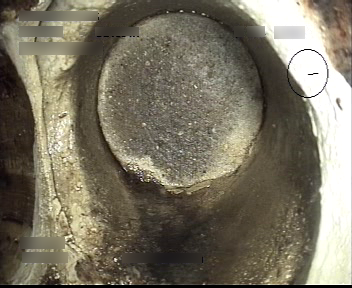} & \includegraphics[width=0.1\textwidth]{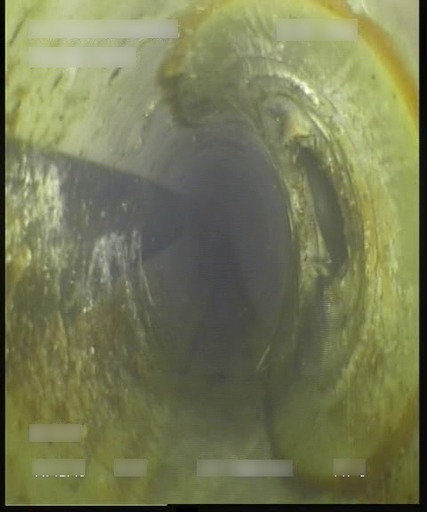} & \includegraphics[width=0.1\textwidth]{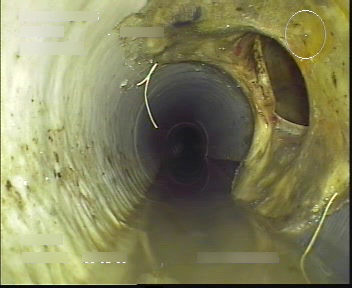} & \includegraphics[width=0.1\textwidth]{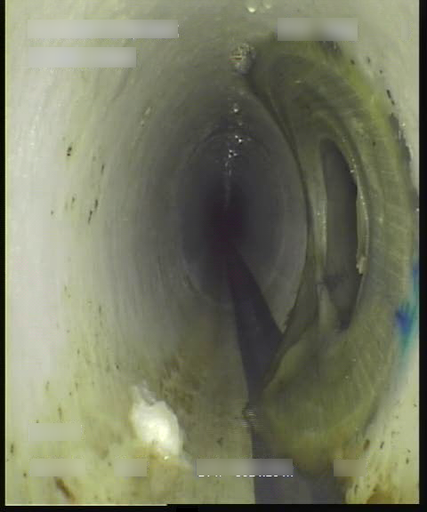} & \includegraphics[width=0.1\textwidth]{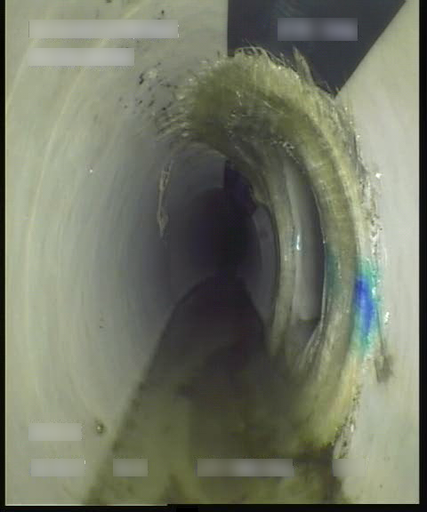} & \includegraphics[width=0.1\textwidth]{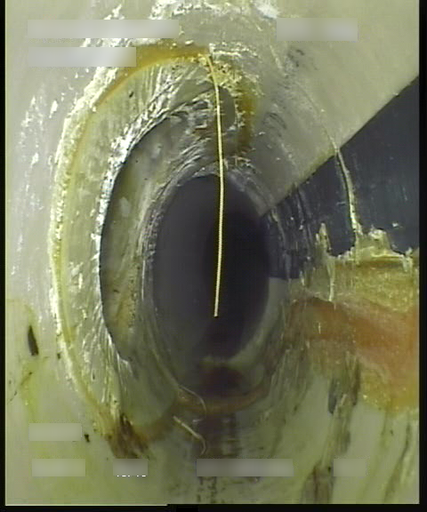} \\
\textbf{RO} & \includegraphics[width=0.1\textwidth]{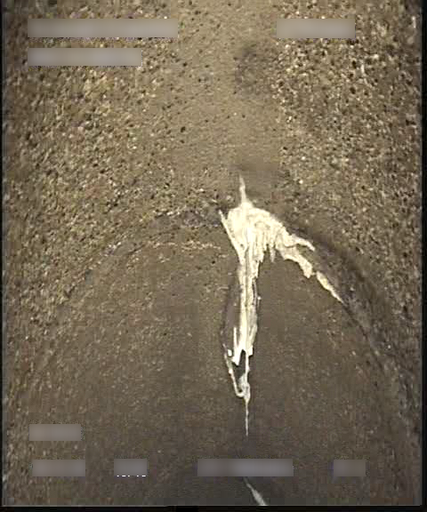} & \includegraphics[width=0.1\textwidth]{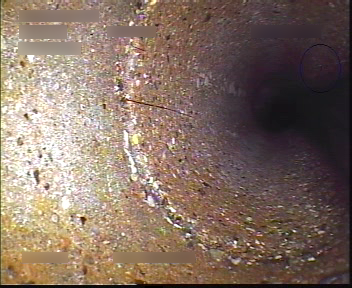} & \includegraphics[width=0.1\textwidth]{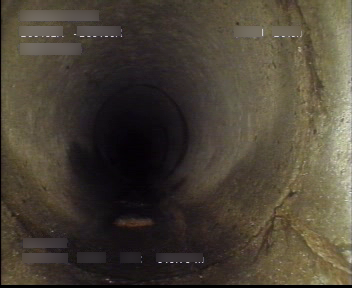} & \includegraphics[width=0.1\textwidth]{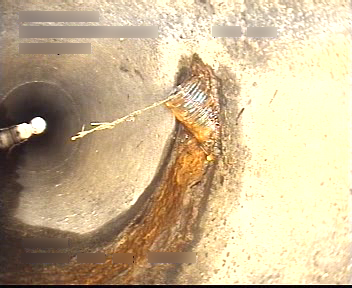} & \includegraphics[width=0.1\textwidth]{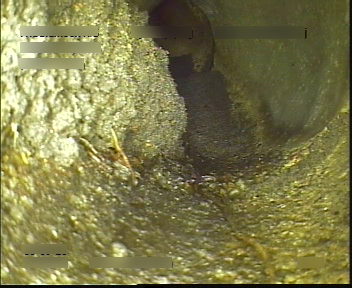} & \includegraphics[width=0.1\textwidth]{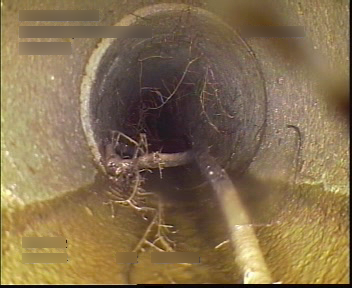} & \includegraphics[width=0.1\textwidth]{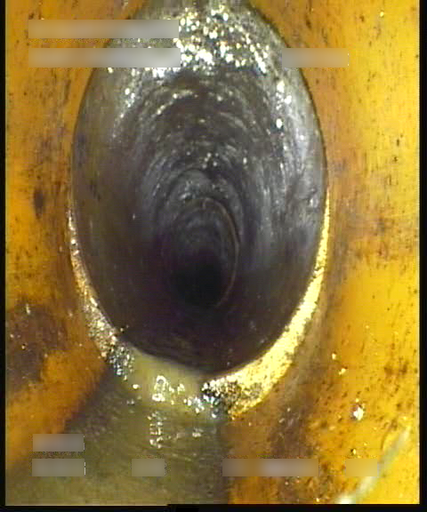} & \includegraphics[width=0.1\textwidth]{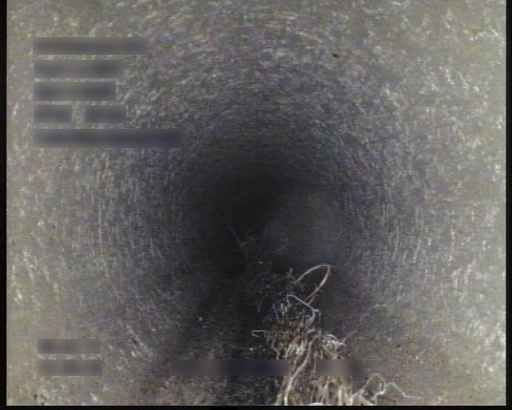} \\
\textbf{IN} & \includegraphics[width=0.1\textwidth]{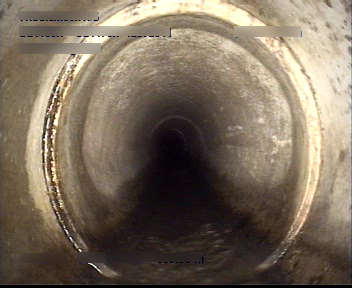} & \includegraphics[width=0.1\textwidth]{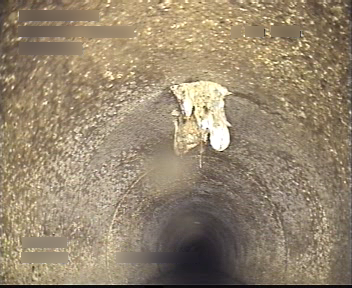} & \includegraphics[width=0.1\textwidth]{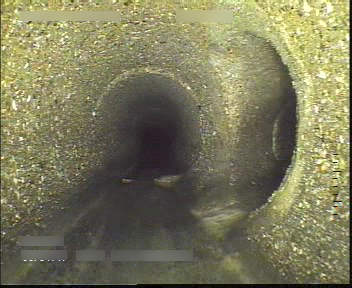} & \includegraphics[width=0.1\textwidth]{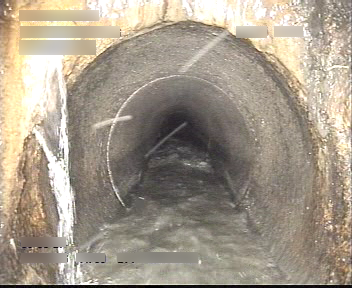} & \includegraphics[width=0.1\textwidth]{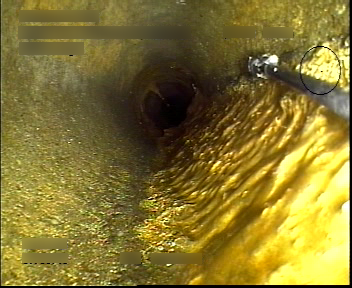} & \includegraphics[width=0.1\textwidth]{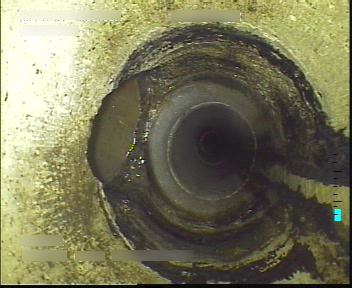} & \includegraphics[width=0.1\textwidth]{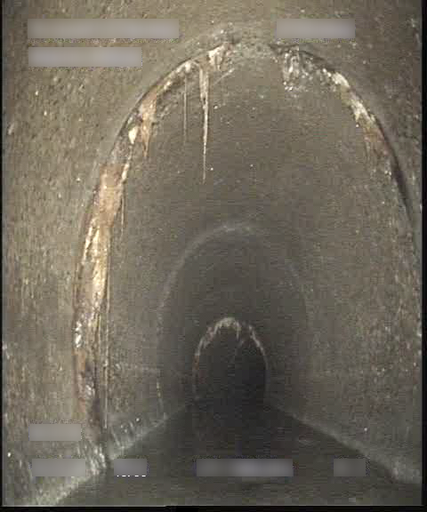} & \includegraphics[width=0.1\textwidth]{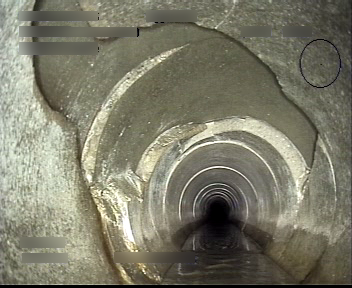} \\
\textbf{PF} & \includegraphics[width=0.1\textwidth]{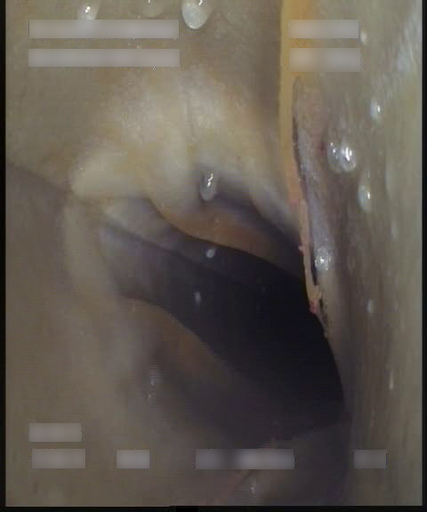} & \includegraphics[width=0.1\textwidth]{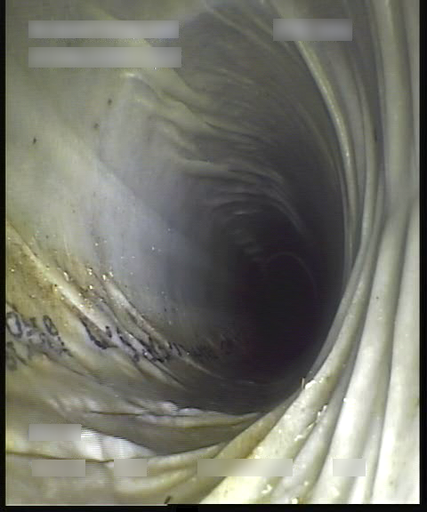} & \includegraphics[width=0.1\textwidth]{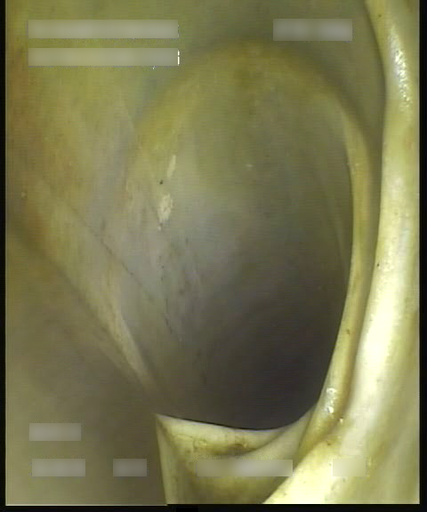} & \includegraphics[width=0.1\textwidth]{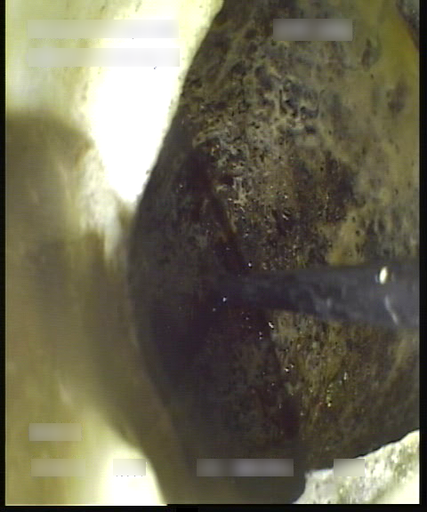} & \includegraphics[width=0.1\textwidth]{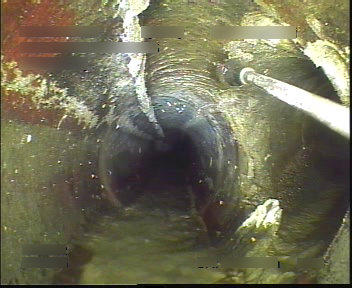} & \includegraphics[width=0.1\textwidth]{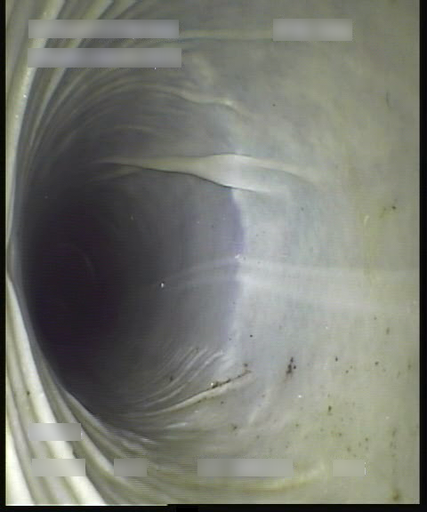} & \includegraphics[width=0.1\textwidth]{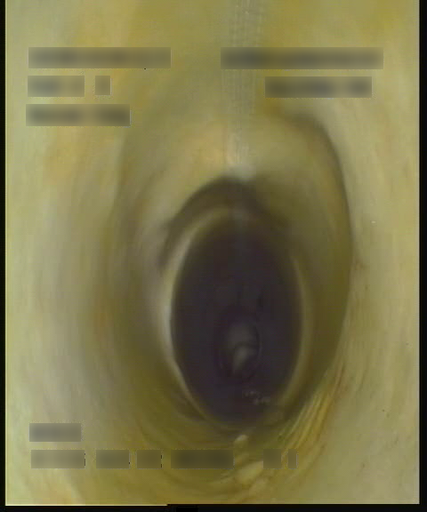} & \includegraphics[width=0.1\textwidth]{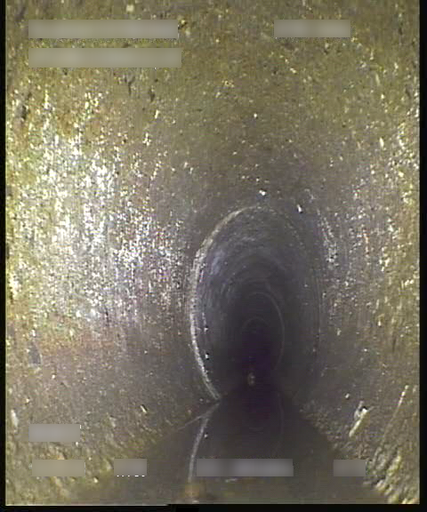} \\
\textbf{FO} & \includegraphics[width=0.1\textwidth]{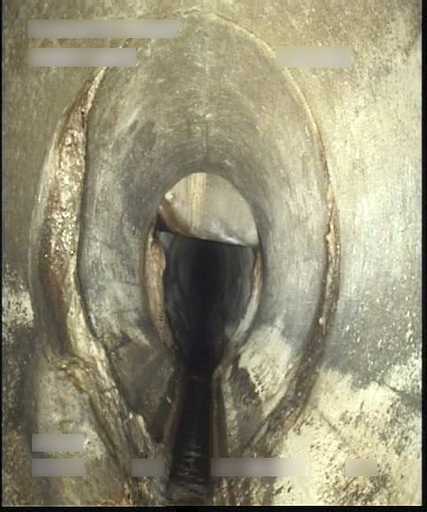} & \includegraphics[width=0.1\textwidth]{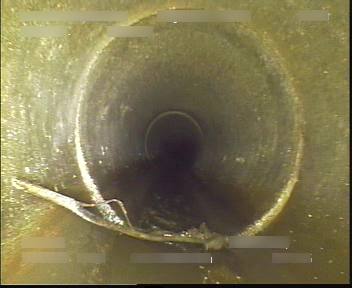} & \includegraphics[width=0.1\textwidth]{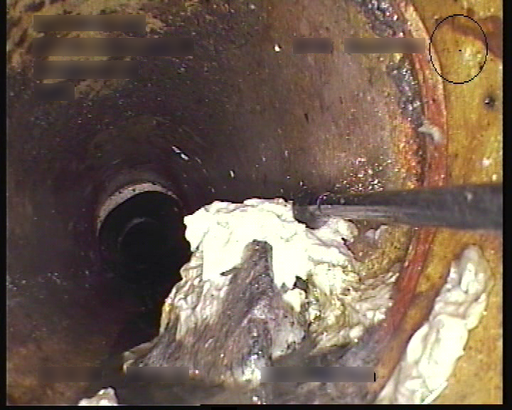} & \includegraphics[width=0.1\textwidth]{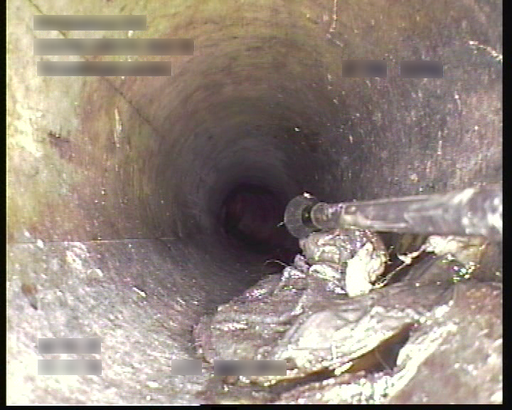} & \includegraphics[width=0.1\textwidth]{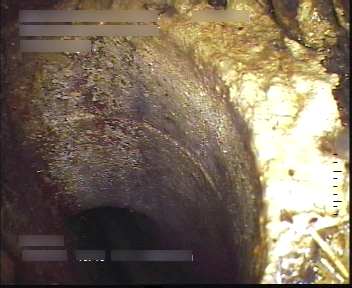} & \includegraphics[width=0.1\textwidth]{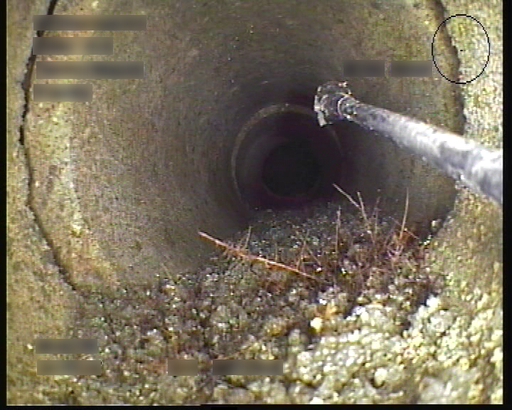} & \includegraphics[width=0.1\textwidth]{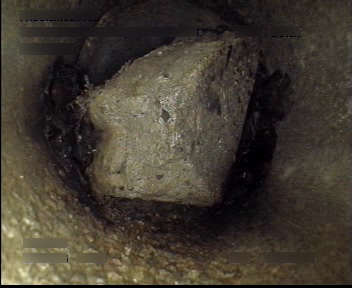} & \includegraphics[width=0.1\textwidth]{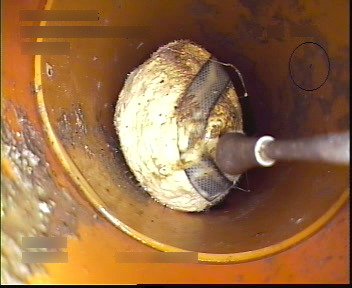} \\
\textbf{BE} & \includegraphics[width=0.1\textwidth]{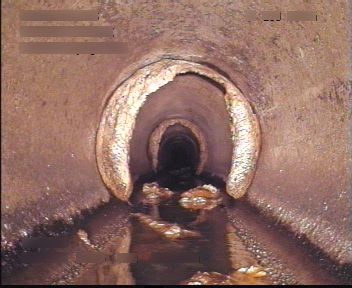} & \includegraphics[width=0.1\textwidth]{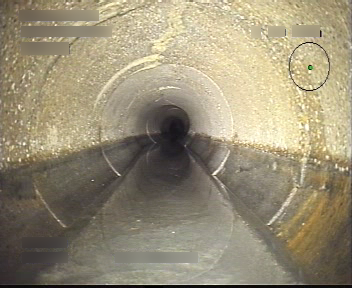} & \includegraphics[width=0.1\textwidth]{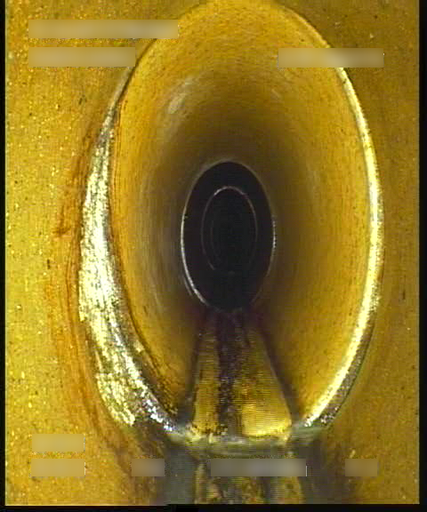} & \includegraphics[width=0.1\textwidth]{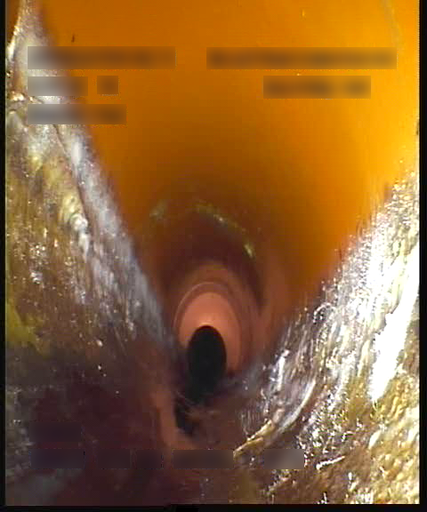} & \includegraphics[width=0.1\textwidth]{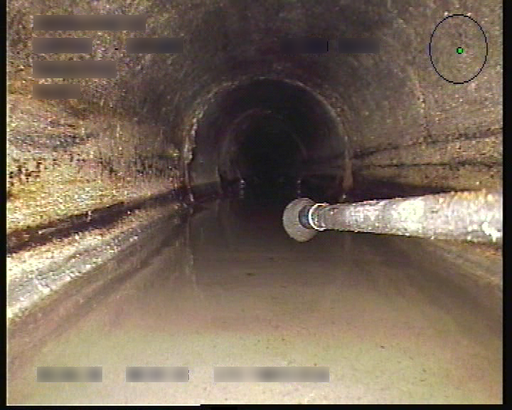} & \includegraphics[width=0.1\textwidth]{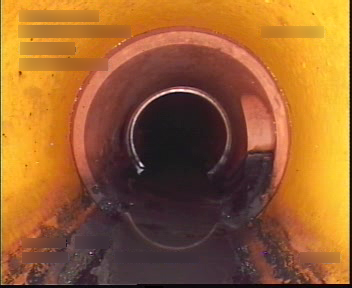} & \includegraphics[width=0.1\textwidth]{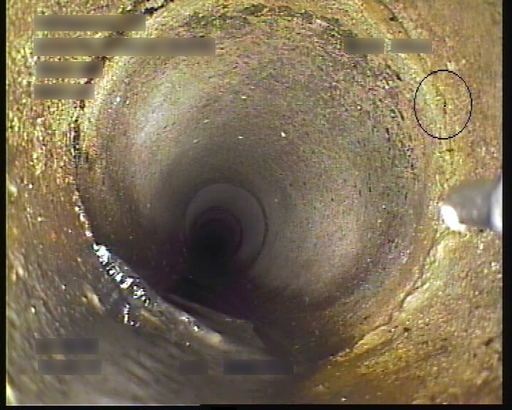} & \includegraphics[width=0.1\textwidth]{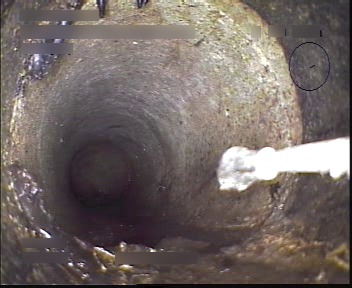} \\
\textbf{IS} & \includegraphics[width=0.1\textwidth]{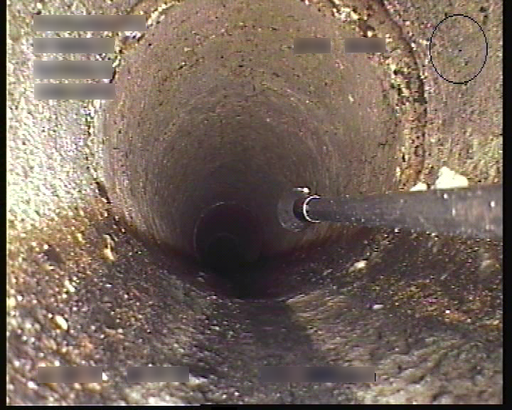} & \includegraphics[width=0.1\textwidth]{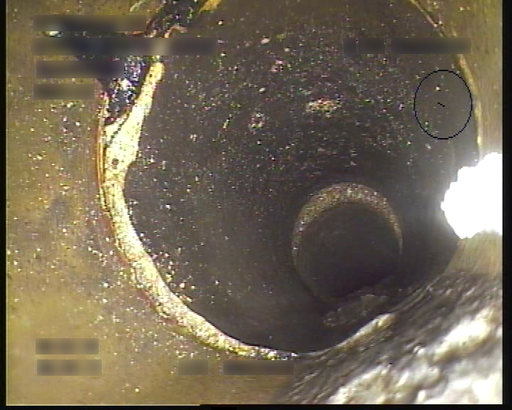} & \includegraphics[width=0.1\textwidth]{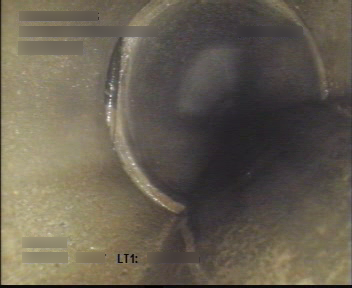} & \includegraphics[width=0.1\textwidth]{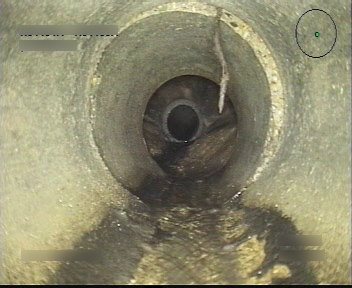} & \includegraphics[width=0.1\textwidth]{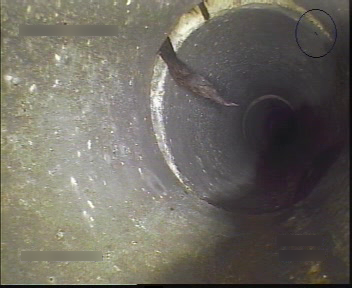} & \includegraphics[width=0.1\textwidth]{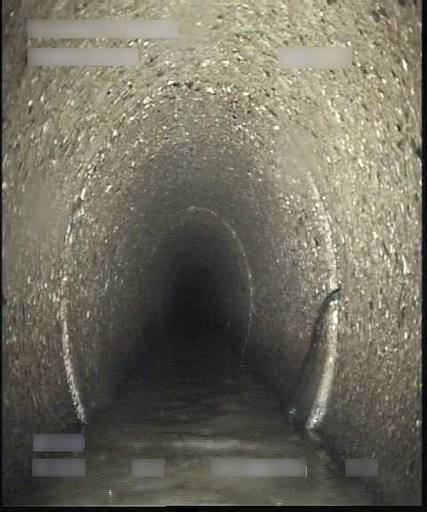} & \includegraphics[width=0.1\textwidth]{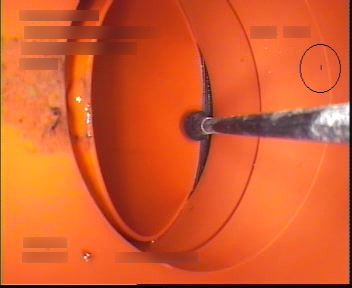} & \includegraphics[width=0.1\textwidth]{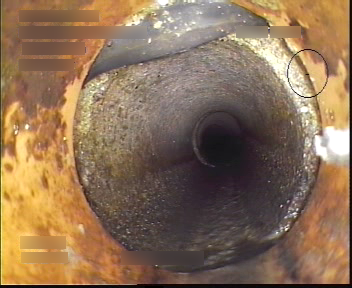} \\
\textbf{DE} & \includegraphics[width=0.1\textwidth]{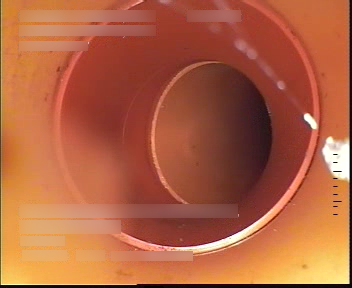} & \includegraphics[width=0.1\textwidth]{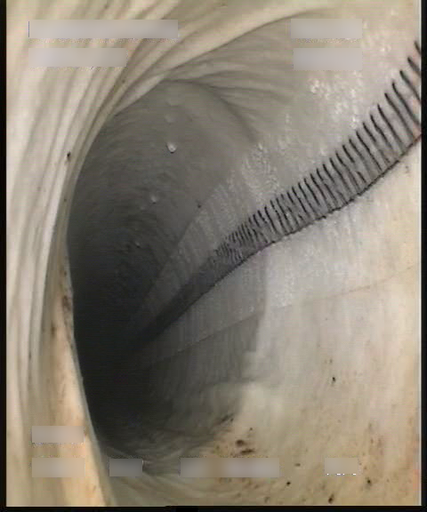} & \includegraphics[width=0.1\textwidth]{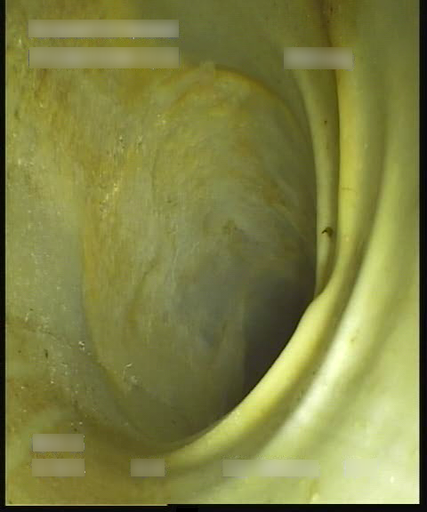} & \includegraphics[width=0.1\textwidth]{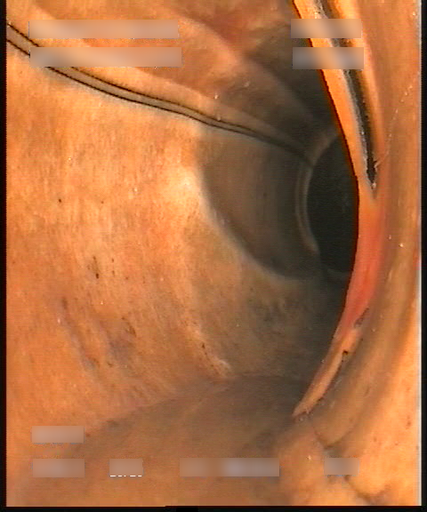} & \includegraphics[width=0.1\textwidth]{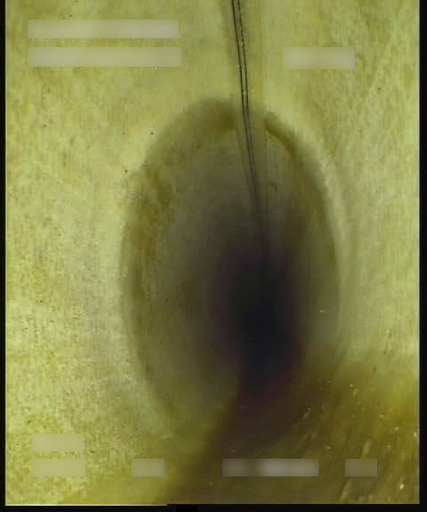} & \includegraphics[width=0.1\textwidth]{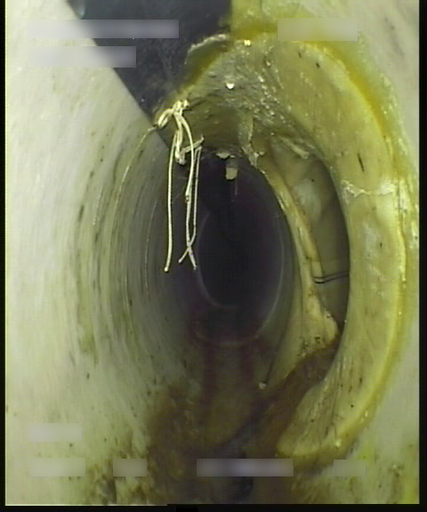} & \includegraphics[width=0.1\textwidth]{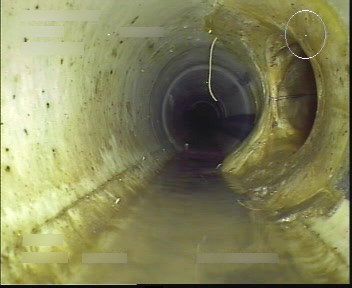} & \includegraphics[width=0.1\textwidth]{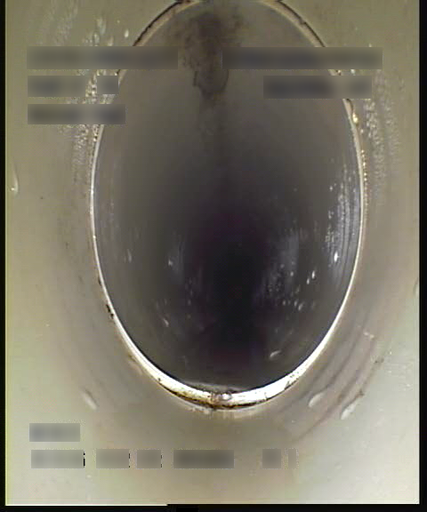} \\
\end{tabular}}
\caption{Sample images by defect type from train and validation sets.}
\end{table}

\end{document}